\documentclass[11pt]{article}
\usepackage{amsmath}
\usepackage{amssymb}
\usepackage[utf8]{inputenc}
\usepackage[margin=1in]{geometry}
\usepackage{graphicx}
\usepackage{caption}
\usepackage{siunitx}
\usepackage{float}
\usepackage{booktabs}
\usepackage[T1]{fontenc}
\usepackage{multirow}
\usepackage{authblk}
\usepackage{natbib}
\usepackage{url}
\usepackage{longtable}

\newcommand{\SIadj}[2]{\SI[number-unit-product=\text{-}]{#1}{#2}}
\newcommand{\SIadjrange}[3]{\SIrange[number-unit-product={\text{-}}]{#1}{#2}{#3}}

\title{Flow matching for Sentinel-2 super-resolution: implementation, application, and implications}
\author[a,*]{Dakota Hester}
\author[a,*]{Vitor S. Martins}
\author[b,a]{Lucas B. Ferreira}
\author[a]{Thainara M. A. Lima}
\author[a]{Juliana A. Araújo}

\affil[a]{Department of Agricultural and Biological Engineering, Mississippi State University}
\affil[b]{North Mississippi Research and Extension Center, Mississippi State University}
\affil[*]{Corresponding authors: \texttt{dh2306@msstate.edu}, \texttt{vmartins@abe.msstate.edu}}
\date{}

\begin{document}

\maketitle

\begin{abstract}
Developing robust techniques for super-resolution of satellite imagery involves navigating commonly observed trade-offs between spectral fidelity and perceptual quality.
In this work, we introduce a flow matching model for 4\(\times\) super-resolution of \SIadj{10}{\meter} Sentinel-2 visible and near-infrared bands over the conterminous United States (CONUS) using a dataset of 120,851 \SIadj{10}{\meter} Sentinel-2 and \SIadj{2.5}{\meter} resampled NAIP imagery pairs acquired on the same day.
Our results showed that the flow matching model outperformed diffusion and Real-ESRGAN models in pixel-wise accuracy in a single sampling step using the Euler method,  closely approaching the performance of a dedicated supervised regression baseline.
When evaluated with a second-order Midpoint solver, our model generated perceptually realistic super-resolved imagery in only 20 sampling steps, effectively navigating the perception-distortion trade-off at inference time without retraining.
We used this model to produce a super-resolved \SIadj{2.5}{\meter} 4-band CONUS imagery product derived from 2025 \SIadj{10}{\meter} Sentinel-2 annual composites, consisting of over 1.58 trillion pixels.
We further evaluated the use of super-resolved data on a land cover classification task using semantic segmentation models.
Finally, we generated a yearly \SIadj{2.5}{\meter} land cover product for the Chesapeake Bay watershed for 2020--2025.
An accuracy assessment against 25,000 ground truth points revealed an overall accuracy of 89.11\%  for the 2021 and 2022 land cover data.
We conclude that flow matching is an effective generative modeling approach for super-resolution of Sentinel-2 imagery compared to diffusion and Generative Adversarial Network-based methods, and has strong implications for expanding access to high-resolution imagery for geospatial applications that demand fine spatial detail.

\end{abstract}


\section{Introduction}\label{sec:introduction}

The use of deep learning super-resolution models for producing synthetic high spatial resolution imagery from medium resolution imagery has expanded rapidly in recent years~\citep{wangLightweightRemoteSensing2025,wangRobustRemoteSensing2024,wangRemoteSensingImage2022}.
Many of these approaches utilize generative adversarial networks (GANs), where a generator network is trained to map coarse-resolution inputs to high-resolution outputs using both pixel-wise loss functions (e.g., \(\mathcal{L}_{1}\) or \(\mathcal{L}_{2}\) loss) and feedback from a discriminator network, which is simultaneously trained to distinguish real high-resolution images from those produced by the generator network~\citep{goodfellowGenerativeAdversarialNets2014, ledigPhotoRealisticSingleImage2017,gongEnlightenGANSuperResolution2021}.
For instance, \citet{jiaMultiattentionGenerativeAdversarial2022} used a GAN framework for 4\(\times\) super-resolution of aerial imagery, outperforming non-adversarial CNN-based super-resolution models.
\citet{zhuQISGANLightweightAdversarial2023} incorporated neural implicit representations into a GAN-based adversarial training framework for hyperspectral image super-resolution.
\citet{phamSpatialResolutionEnhancement2021} used an Enhanced Super-Resolution GAN (ESRGAN) architecture~\citep{wangESRGANEnhancedSuperResolution2018} to super-resolve \SIadj{30}{\meter} Landsat-8 visible and near-infrared (NIR) bands to \SIadj{10}{\meter} resolution using temporally aligned Sentinel-2 images as target outputs during training.
\citet{crivellariSuperresolutionGANsUpscaling2023} applied an SRGAN architecture~\citep{ledigPhotoRealisticSingleImage2017} to super-resolve the \SIadj{10}{\meter} Sentinel-2 bands to \SIadj{1}{\meter} using GaoFen-2 imagery as ground truth, demonstrating improved land cover classification performance when using the super-resolved imagery compared to simple bicubic upsampling.

However, GAN-based super-resolution approaches suffer from several fundamental drawbacks, the most commonly cited of which is the instability of the adversarial training process~\citep{arjovskyWassersteinGenerativeAdversarial2017}.
This adversarial process makes robust training inherently difficult, frequently leading to mode collapse, where the generator model only learns to capture a small portion of the target distribution accurately, resulting in generative models that fail to produce photorealistic samples~\citep{salimansImprovedTechniquesTraining2016}.
Further, the convolutional nature of discriminator networks leads to instances where the discriminator is blind to certain types of artifacts in generated images, incentivizing the generator network to produce images laden with these artifacts~\citep{odenaDeconvolutionCheckerboardArtifacts2016, liangDetailsArtifactsLocally2022}.
To address this, many super-resolution GANs incorporate a perceptual loss function during training, where a pre-trained feature extraction network is used to provide additional feedback to the generator network regarding the perceptual similarity between super-resolved imagery and their ground truth counterparts~\citep{ledigPhotoRealisticSingleImage2017,wangESRGANEnhancedSuperResolution2018}.
Nevertheless, these perceptual losses introduce a different class of distortions into generated imagery, as they prioritize similarity between extracted features over pixel-wise accuracy, leading to images with photorealistic textures but poor spectral fidelity.
\citet{blauPerceptionDistortionTradeoff2018} first formalized this inherent trade-off between perceptual quality and distortion (i.e., pixel-wise accuracy) in generative image modeling tasks, showing that GAN-based techniques can model this behavior by adjusting the weighting between the pixel-wise and adversarial or perceptual loss terms during training, but cannot simultaneously optimize for both perceptual quality and low distortion.
\citet{aybarRadiometricallySpatiallyConsistent2026} emphasized that this trade-off is particularly relevant for Sentinel-2 super-resolution and argued that GAN-based super-resolution models disproportionately prioritize perceptual quality over spectral reliability.
This bias towards perceptual quality yields super-resolved imagery that is visually pleasing but often poorly suited for downstream analytical tasks.
Conversely, for analyses where spatial features are more critical than spectral fidelity (e.g., road network extraction), perceptual quality may be more important.
However, developing models that can flexibly adapt to different points along this trade-off at inference time remains a challenge.

Motivated by the drawbacks of GAN frameworks for generative image modeling tasks, research efforts have recently shifted towards iterative denoising models based on diffusion processes~\citep{croitoruDiffusionModelsVision2023}.
Unlike GANs, these architectures resemble semantic segmentation and image-to-image translation models, as they are a single network that transforms an input image to an output image,  and are often trained using a single pixel-wise loss function without any additional adversarial feedback or perceptual loss terms.
The training process for diffusion models relies on a multi-step iterative denoising process, where small amounts of noise are added to images over a series of steps, and the model is trained to reverse this noising process.
At inference time, the model can be used to generate new images by starting from pure noise and iteratively denoising the image over a series of steps to produce a final output image~\citep{sohl-dicksteinDeepUnsupervisedLearning2015,hoDenoisingDiffusionProbabilistic2020,songDenoisingDiffusionImplicit2022}.
This process can easily be extended to super-resolution tasks by reframing the task as a conditional image generation problem, where the diffusion model follows a similar training routine but incorporates some form of conditioning mechanism to guide the denoising process toward a high-resolution output that corresponds to a given low-resolution conditioning input~\citep{sahariaImageSuperResolutionIterative2021,liSRDiffSingleImage2022,gaoImplicitDiffusionModels2023}.
This trend away from GANs towards diffusion-based approaches has recently emerged in generative remote sensing applications as well~\citep{liuDiffusionModelsMeet2024}.
\citet{xiaoEDiffSREfficientDiffusion2024} developed a custom U-Net-like denoising model that utilized enhanced conditioning mechanisms for aerial imagery super-resolution, showing that diffusion-based super-resolution models can outperform GAN-based approaches in image quality and spectral fidelity.
\citet{wangSemanticGuidedLarge2025} used a diffusion-based framework for \(16\times\) super-resolution of Sentinel-2 imagery using vector map data as additional conditioning information, demonstrating that the diffusion-based approach outperformed GAN-based super-resolution models and reduced hallucinations (i.e., features that appear realistic but are absent in the true high-resolution imagery).
\citet{chenSpectralCascadedDiffusionModel2024} used diffusion for spectral super-resolution, enabling the generation of synthetic hyperspectral imagery with dozens of bands from 3-band RGB inputs.
\citet{mengConditionalDiffusionModel2024} used an aggressive noise schedule for faster sampling (i.e., inference) for super-resolution of aerial imagery in 20 denoising steps (compared to the typical 100+ steps used in most diffusion models) while maintaining high image quality.

Yet, despite the relative stability of training (due to the simple pixel-wise loss function) and high-quality outputs produced by diffusion models, they require many denoising steps at inference time to produce high-quality images.
This leads to long inference times compared to GANs, which only require a single forward pass through the generator network to produce an output image~\citep{miaoResearchCrossSensorRemote2025,songScoreBasedGenerativeModeling2021}.
One promising approach that has recently emerged as a means of reducing the computational burden of generative models is the use of continuous normalizing flows as an alternative to diffusion processes for training generative image models.
Instead of modeling the denoising process as a stochastic Markov process, flow matching simplifies the process to a straightforward ordinary differential equation (ODE) that uses a linear flow field, defined using the optimal transport between the data distribution and a simple Gaussian noise distribution.
Thus, the model is trained to predict this ODE flow field, and at inference time, the model can produce high-quality images from noise using a simple numerical ODE solver (e.g., Euler's method) to integrate the learned flow field over a series of steps.
The use of flow matching-based generative models has been shown to facilitate the generation of high-quality synthetic images using substantially fewer forward passes through the denoising network (e.g., 10 forward passes instead of 100)~\citep{lipmanFlowMatchingGenerative2023,pooladianMultisampleFlowMatching2023}.
Yet, no prior work has rigorously explored the use of flow matching for super-resolution of Sentinel-2 imagery, nor its utility for downstream remote sensing data analysis tasks.

This study presents the first national-scale implementation of a flow matching framework for Sentinel-2 super-resolution and provides a rigorous evaluation of the resulting imagery's utility for downstream land cover classification.
Following the super-resolution via repeated refinement (SR3) framework for super-resolution with diffusion models~\citep{sahariaImageSuperResolutionIterative2021}, we trained a flow matching model to generate \SIadj{2.5}{\meter} resolution synthetic visible and NIR imagery from \SIadj{10}{\meter} Sentinel-2 inputs using a large dataset of Sentinel-2 and cross-calibrated aerial imagery from the USDA NAIP archive.
 We demonstrated that simple ODE solvers that use the Euler and Midpoint methods are more than sufficient to produce high-quality super-resolved imagery, outperforming diffusion-based approaches such as DDPM~\citep{hoDenoisingDiffusionProbabilistic2020} and DDIM~\citep{songDenoisingDiffusionImplicit2022} in both image quality metrics and inference time, while providing more spectrally accurate results compared to a Real-ESRGAN super-resolution model and closely approaching the pixel-wise accuracy of a dedicated supervised U-Net regression baseline.
Using this approach, we produced a synthetic \SIadj{2.5}{\meter} visible/NIR imagery product over the entirety of the conterminous United States (CONUS) for the year 2025, providing a new high-resolution imagery data product for the remote sensing community.
Further, we evaluated the utility of our synthetic \SIadj{2.5}{\meter} imagery for downstream land cover classification tasks using semantic segmentation models.
This application revealed that the perception-distortion trade-off is highly relevant for downstream analysis, as the perceptually superior synthetic imagery led to worse land cover classification accuracy compared to more spectrally accurate super-resolved imagery.
These findings are crucial, as the super-resolved imagery produced by our flow matching framework demonstrated suitability for reliable land cover classification performance in a single inference step, while also being able to produce perceptually superior imagery by simply increasing the number of ODE solver steps at inference time.
Using the best-performing segmentation model evaluated, we produced and validated an annual \SIadj{2.5}{\meter} land cover data product for the Chesapeake Bay watershed for 2020--2025, providing a temporally dense, high-resolution land cover dataset for the region.
In summary, this study introduces a novel application of flow matching models for Sentinel-2 super-resolution using cross-sensor training data and provides a rigorous evaluation of the implications of such a product for downstream remote sensing data tasks, such as land cover mapping.

\section{Data sources}\label{sec:data}

\subsection{Sentinel-2 Level-2A data}

The European Space Agency's (ESA) Sentinel-2 satellite constellation provides publicly available multispectral imagery with a 10-day revisit period across three satellites, giving the constellation an effective temporal resolution of 5 days for most terrestrial sites across the globe~\citep{druschSentinel2ESAsOptical2012}.
The first satellite, Sentinel-2A, was launched on June 23, 2015, followed by Sentinel-2B on March 7, 2017, and Sentinel-2C on September 4, 2024.
The Multispectral Instrument aboard each satellite records visible and NIR optical reflectance data across 13 spectral bands with 12-bit radiometric resolution, with bands 2 (blue, \SI{490}{\nano \meter}), 3 (green, \SI{560}{\nano \meter}), 4 (red, \SI{665}{\nano \meter}), and 8 (NIR, \SI{842}{\nano \meter}) acquired at a spatial resolution of \SI{10}{\meter}.
Because the Sentinel-2 constellation provides temporally dense imagery at high spatial resolution, it has seen broad use across a wide variety of applications where high resolution, temporally dense imagery at regional-to-global scale is needed~\citep{brownDynamicWorldRealtime2022,blickensdorferMappingCropTypes2022,jiaMappingGlobalDistribution2023}.

For the super-resolution model input imagery, we used the visible and NIR bands of Sentinel-2 Level-2A surface reflectance products at \SIadj{10}{\meter} resolution.
Sen2Cor is ESA's official Level-2A processor for Sentinel-2 data, providing atmospherically corrected surface reflectance values as well as a Scene Classification Layer (SCL) that provides quality information for each pixel at a \SIadj{20}{\meter} resolution~\citep{main-knornSen2CorSentinel22017}.
We used SCL to filter out images in which the surface is obscured by cloud cover, cloud shadows, or terrestrial shadows, or which are otherwise of low quality due to oversaturation or data gaps.
The \SIadj{10}{\meter} Sentinel-2 surface reflectance data serve as the primary input to the super-resolution models developed in this study.
We used Microsoft Planetary Computer's \texttt{sentinel-2-l2a} collection to access Sen2Cor-processed Sentinel-2 surface reflectance data~\citep{microsoftopensourceMicrosoftPlanetaryComputerOctober2022}.

\subsection{NAIP aerial imagery}

The target high-resolution imagery for this study is sourced from the USDA's National Agriculture Imagery Program (NAIP)~\citep{earthresourcesobservationandscienceeroscenterNationalAgricultureImagery2017}.
NAIP provides an archive of high spatial resolution (\SIadjrange{0.3}{0.6}{\meter}) multispectral aerial imagery across CONUS during the peak agricultural growing season (June to September).
Each state is typically imaged every 2--3 years, with imagery collection starting in 2003 and continuing to the present.
Imagery is collected using crewed aircraft and is orthorectified to account for terrain and geometric distortions.
NAIP imagery is mostly cloud-free due to strict quality control measures during acquisition, making it a high-quality data source for numerous downstream applications.
Consequently, NAIP imagery has a long history of use as source imagery for a variety of terrestrial remote sensing applications over CONUS requiring high spatial resolution imagery~\citep{maxwellLandCoverClassification2017,airesNationalscaleOpenCattle2026,bhattComparisonHighresolutionNAIP2023,martinsDigitalMappingStructural2021,martinsExploringMultiscaleObjectbased2020}.
More recently, NAIP imagery has also seen use as a target high-resolution dataset for training and evaluating Sentinel-2 super-resolution models~\citep{aybarComprehensiveBenchmarkOptical2024,aybarSEN2NAIPLargescaleDataset2024,aybarRadiometricallySpatiallyConsistent2026,woltersZoomingOutZooming2023}.
For this study, we used 4-band NAIP imagery (Red, Green, Blue, NIR) acquired between 2015 and 2023 to match the temporal coverage of Sentinel-2 data.
As with the Sentinel-2 data, we used Microsoft Planetary Computer's \texttt{naip} collection to query and download NAIP imagery.

\subsection{Chesapeake Bay Land Use/Land Cover Dataset}

To demonstrate the utility of the super-resolved imagery generated by the super-resolution models, we implemented a land cover mapping task.
We utilized the Chesapeake Bay Land Use/Land Cover Dataset (CBLC), a land use and land cover dataset covering the Chesapeake Bay watershed at \SIadj{1}{\meter} spatial resolution, as a reference dataset.
The 2024 edition dataset provides a 56-class land use product derived from NAIP imagery across three temporal periods: 2013/2014, 2017/2018, and 2021/2022, the specific year within each period being a function of the NAIP imagery acquisition schedule for a given state within the watershed~\citep{claggettChesapeakeBayLand2025,mcdonaldChesapeakeBayLand2025}.
As the dataset is derived from NAIP imagery, it is an excellent choice of ground truth land cover labels, as it is already temporally aligned with the NAIP imagery used in this study.
For this study, we used only the 2021/2022 land cover labels, as this data has undergone a formal accuracy assessment  and constitutes the most up-to-date edition of the product at the time the study was performed.

\newcommand{\gennetwork}{f_\theta^G}
\newcommand{\discnetwork}{f_\theta^D}

\section{Methodology}\label{sec:methodology}

\begin{figure}[H]
    \centering
    \includegraphics[width=\textwidth]{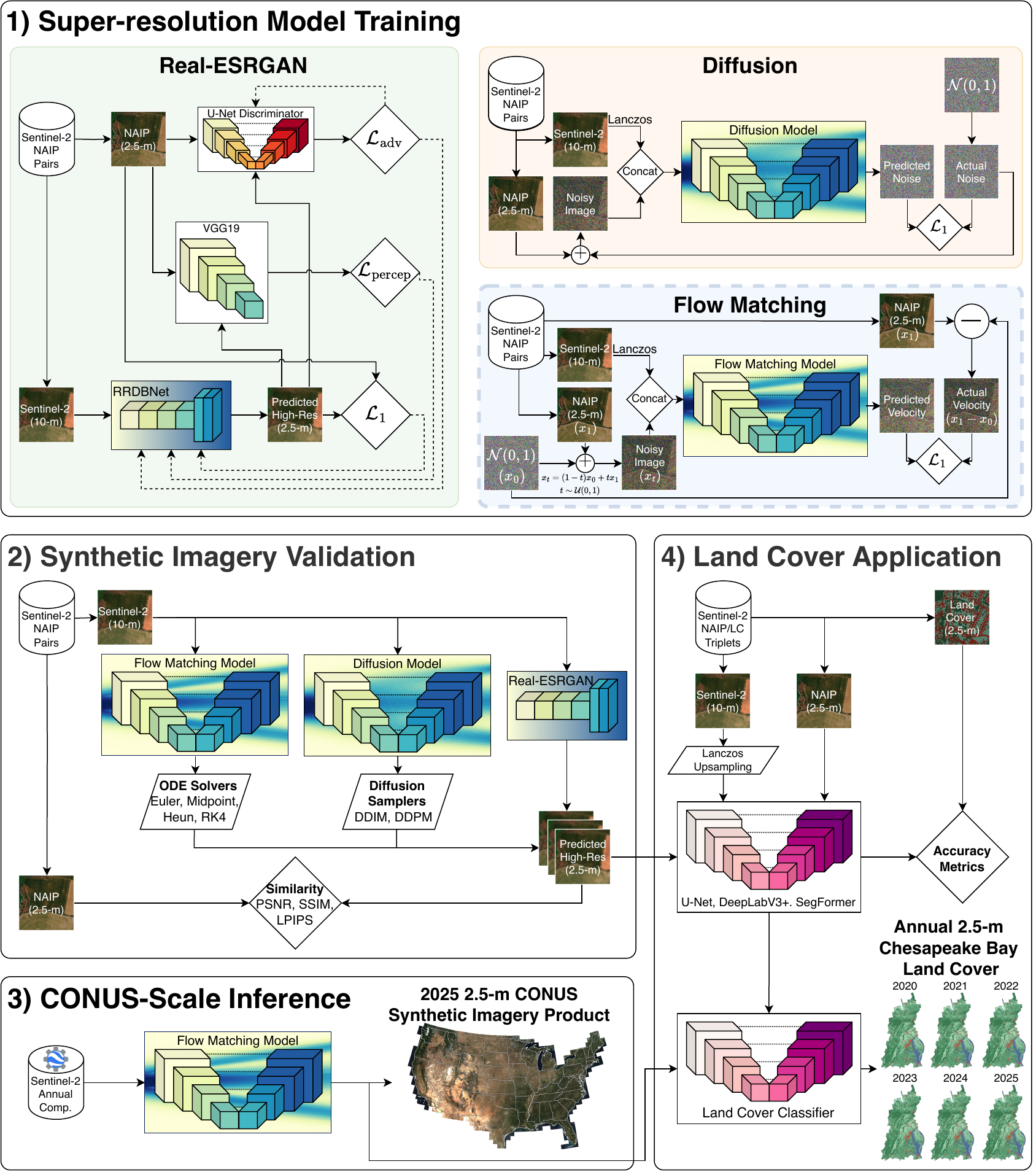}

    \begingroup
    \captionsetup{
    }
    \caption{A high-level overview of the methodology used in this study.}\label{fig:methodology_flowchart}
    \endgroup
\end{figure}

Our methodology utilized Sentinel-2 and cross-calibrated NAIP imagery pairs to train several super-resolution models (Section~\ref{sec:methodology:preprocessing:s2naip}) and a separate Sentinel-2/NAIP/land cover triplet dataset to train land cover classification models over the Chesapeake Bay watershed (Section~\ref{sec:methodology:preprocessing:cbp}).
 We trained and evaluated several super-resolution modeling frameworks, including our proposed flow matching model, diffusion models, Real-ESRGAN, and a supervised L1 regression U-Net baseline (Section~\ref{sec:methodology:sr}).
Both pixel-wise and perceptual similarity metrics were used to evaluate the super-resolution models (Section~\ref{sec:methodology:sr_evaluation}), and the best-performing model and configuration were used to produce a super-resolved \SIadj{2.5}{\meter} imagery product over CONUS for 2025 (Section~\ref{sec:methodology:conus_sr}).
We then evaluated the utility of the super-resolution models as part of a land cover classification pipeline, ultimately resulting in the creation of annual \SIadj{2.5}{\meter} land cover maps for the Chesapeake Bay watershed between 2020 and 2025 (Section~\ref{sec:methodology:lc}), followed by a formal accuracy assessment using the 25k validation points selected in Section~\ref{sec:methodology:preprocessing:cbp}.
We provide a flowchart summarizing the overall methodology in Figure~\ref{fig:methodology_flowchart}.

\subsection{Preprocessing and sampling procedures}\label{sec:methodology:preprocessing}

\subsubsection{Sentinel-2/NAIP pair dataset creation for super-resolution model training and evaluation}\label{sec:methodology:preprocessing:s2naip}

\begin{figure}[H]
    \centering
    \includegraphics[width=\textwidth]{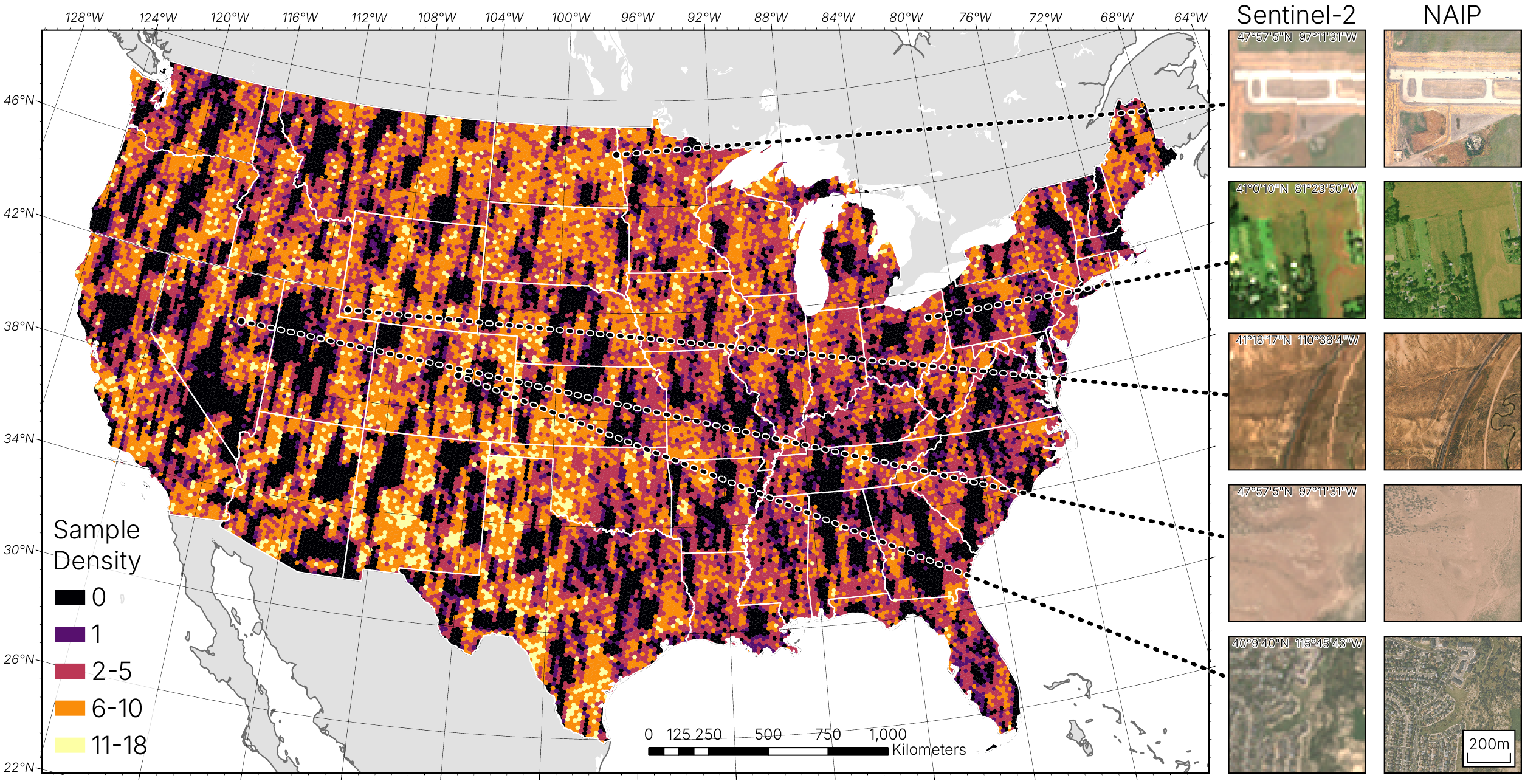}
    \caption{Spatial distribution of Sentinel-2 and NAIP image pairs used for training and evaluating super-resolution models (left) and examples of Sentinel-2 surface reflectance and calibrated NAIP imagery (right).}\label{fig:s2naip_samples}
\end{figure}

To train and evaluate the super-resolution models, we curated a dataset of Sentinel-2 and NAIP imagery pairs across CONUS.
First, we identified regions where both Sentinel-2 and NAIP images were collected on the same date to minimize discrepancies arising from temporal changes.
Using Microsoft Planetary Computer's STAC API, we queried the footprints of Sentinel-2 Level-2A surface reflectance products and NAIP orthophotos collected between 2015 and 2023, and intersected temporally aligned footprints to produce a broad set of candidate regions.
From this set of candidate regions, we randomly sampled 250,000 locations to serve as centroids for 640\(\times\)\SIadj{640}{\meter} square polygons from which chips of imagery would be extracted.
A minimum distance of 2 km was enforced between sampled locations to avoid data leakage or interdependence between samples arising from spatial autocorrelation.
We randomly withheld 5\% of the sampled locations for validation, while the remaining 95\% were used for training.
At each sample location, we again queried Microsoft Planetary Computer's STAC API to retrieve the Sentinel-2 SCL for a random valid year (i.e., a year where both Sentinel-2 and NAIP imagery were available on the same date).
If any pixel in a potential sample was classified as saturated/defective (SCL class 1), topographic shadow (SCL class 2), cloud shadows (SCL class 3), unclassified (SCL class 7), or cloud (SCL class 8, 9, 10), we searched for a new Sentinel-2 and NAIP image pair at that location in a different year.
If no other years contained a valid image pair or if no other years were available, we discarded that sample location.
Once a sample had passed this quality check, we downloaded the corresponding Sentinel-2 and NAIP imagery.
The final dataset consisted of 114,835 training samples and 6,016 validation samples for a total of 120,851 Sentinel-2/NAIP image pairs.
The spatial distribution of the sampled locations is shown in Figure~\ref{fig:s2naip_samples}.

We extracted 640\(\times\)\SIadj{640}{\meter} chips from both Sentinel-2 and NAIP imagery samples.
Both images were reprojected to EPSG:5070 (NAD83/CONUS Albers) to ensure consistency across the dataset, with the NAIP imagery being resampled to the desired \SIadj{2.5}{\meter} resolution using Lanczos resampling~\citep{duchonLanczosFilteringOne1979}.
As the NAIP imagery is not a surface reflectance product, we performed a cross-calibration of the raw digital numbers using the Sentinel-2 surface reflectance values as a reference.
First, we downsampled and reprojected the NAIP imagery to match the \SIadj{10}{\meter} resolution and affine transform of the Sentinel-2 imagery chip to align the two images spatially.
Next, we fitted a per-band ordinary least squares linear regression model between the NAIP pixel values and the corresponding Sentinel-2 surface reflectance values on a per-image basis.
We then applied the learned linear transformation to the original \SIadj{2.5}{\meter} NAIP imagery chip to produce estimates of surface reflectance at \SIadj{2.5}{\meter} resolution.
These Sentinel-2 and cross-calibrated NAIP imagery pairs were stored as 16-bit unsigned integer GeoTIFF files for later use during model training and evaluation.
Figure~\ref{fig:s2naip_samples} shows a selection of Sentinel-2 and cross-calibrated NAIP image pairs from the dataset.

\subsubsection{Sentinel-2/NAIP/land cover label triplet dataset creation for super-resolution land cover classification}\label{sec:methodology:preprocessing:cbp}

To train the segmentation models for land cover classification, we curated a dataset of Sentinel-2/NAIP/land cover label triplets.
This process followed a similar procedure to the Sentinel-2/NAIP pair dataset creation process described in Section~\ref{sec:methodology:preprocessing:s2naip}, but with a few key differences.
First, we did not utilize prior years of CBLC data to identify valid sample locations, instead relying solely on the most recent 2021/2022 data for sampling.
This constrained the number of potential Sentinel-2/NAIP concomitant acquisitions since we could not sample from across multiple years.
To account for this, we allowed for a greater degree of temporal mismatch between Sentinel-2 and NAIP acquisitions, allowing for up to a 3-day difference between imagery collection dates when querying for candidate regions.
We followed the same procedure as in Section~\ref{sec:methodology:preprocessing:s2naip} to identify potential sample locations prior to sampling 70,000 locations across the valid regions.
Instead of a 2 km minimum distance between samples, we relaxed this constraint to 1.5 km to allow for a greater density of samples given the smaller spatial extent of the Chesapeake Bay watershed.
We placed 20\% of the sampled locations into a withheld test split, while the remaining 80\% were randomly distributed among 5 folds for cross-validation during model training.
We again enforced the same quality checks as before to ensure that no clouds, cloud shadows, or defective pixels were present, and that all data points in the Sentinel-2 imagery, NAIP imagery, and land cover labels were valid.
In total, we obtained 28,446 valid samples of Sentinel-2/NAIP/land cover label triplets to be used for training and evaluating land cover classification models using super-resolved Sentinel-2 imagery, with 5,689 samples reserved for testing and 22,757 samples used for 5-fold cross-validation.

As with the Sentinel-2/NAIP pair dataset, all imagery/data chips were reprojected to EPSG:5070 (NAD83/CONUS Albers) for consistency; the NAIP imagery was resampled to \SIadj{2.5}{\meter} resolution using Lanczos resampling, while the land cover labels were resampled to \SIadj{2.5}{\meter} resolution using mode resampling.
We chose to reclassify the land use labels from the original CBLC dataset into a simplified set of 5 general land cover classes.
While a 5-class legend is somewhat coarse, many ancillary datasets (e.g., wetland maps, building footprints, road networks, cropland maps, etc.) were used in the original classification process of the data.
As we were primarily concerned with evaluating the utility of super-resolved Sentinel-2 imagery for land cover classification tasks, we opted to simplify the legend so as to remove the dependence on these ancillary datasets and focus solely on spectral and spatial information present in the imagery.
The reclassification scheme is provided in Table S1.

\begin{figure}[H]
    \centering
    \includegraphics[width=\textwidth]{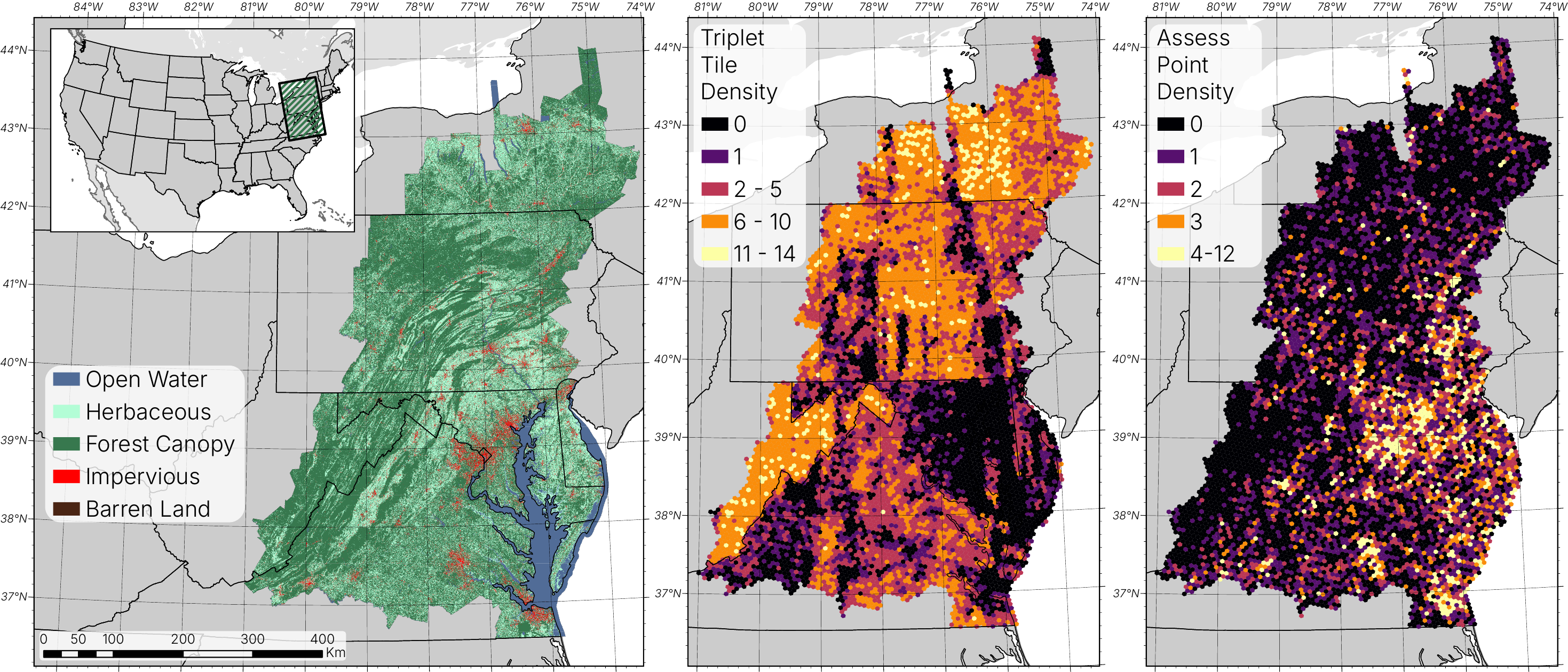}
    \caption{Reclassified CBLC data (left), locations of triplets used for training land cover classification models (middle), and locations of points used for accuracy assessment of annual \SIadj{2.5}{\meter} land cover product (right).}
    \label{fig:cpb_samples}
\end{figure}

To validate the resulting \SIadj{2.5}{\meter} land cover product, we curated a second dataset of 25,000 points across the Chesapeake Bay watershed for an accuracy assessment of the land cover product.
Points were randomly sampled across the region with a minimum distance of 2 km between points and tiles sampled for the Sentinel-2/NAIP/land cover triplet dataset to minimize the effect of spatial autocorrelation between the two datasets and ensure spatial independence of the accuracy assessment points.
At each point, we collected the corresponding ground-truth land cover label from the original CBLC dataset, then reclassified the label according to the scheme in Table S1 to match the legend of the land cover product.
The spatial extent of the CBLC dataset, the density of sampled triplets for land cover classifier training/evaluation, and the density of sampled points for the accuracy assessment dataset are shown in Figure~\ref{fig:cpb_samples}.

\subsubsection{Sentinel-2 annual mean composites}\label{sec:methodology:preprocessing:composites}

To create both the super-resolved CONUS \SIadj{2.5}{\meter} imagery product and the annual \SIadj{2.5}{\meter} land cover maps for the Chesapeake Bay watershed, we opted to use annual mean composites of Sentinel-2 imagery as the input to the super-resolution models instead of single-date imagery.
This decision was motivated by the fact that annual composites are more likely to be cloud-free, and while choosing cloud-free tiles is possible for single-date imagery, this would lead to products that are temporally inconsistent across space and time.
Annual mean composites, on the other hand, provide a consistent snapshot of the average landscape of a region within a given year.
Google Earth Engine was used to generate annual mean composites of Sentinel-2 Level-2A imagery for the year 2025 across CONUS and for 2020--2025 across the Chesapeake Bay watershed~\citep{gorelickGoogleEarthEngine2017}.
We used the SCL layer of the Sentinel-2 Level-2A product to remove snow/ice pixels and the Cloud Score+ quality assessment product provided by Google Earth Engine to mask out pixels that were predicted as being less than 65\% clear~\citep{pasquarellaComprehensiveQualityAssessment2023} during the compositing process.

\subsection{ Super-resolution modeling frameworks}\label{sec:methodology:sr}

\subsubsection{ Supervised \(\mathcal{L}_1\) Regression U-Net}\label{sec:methodology:sr:unet_baseline}

Perhaps the most straightforward method for model-driven super-resolution is to approach the problem as a supervised regression task.
Under this formulation, a neural network \(f_\theta\) is trained to learn a deterministic mapping from a low-resolution input \(x\) to a high-resolution output \(y\) by minimizing a pixel-wise distance metric between the predicted output \(f_\theta(x)\) and the ground truth \(y\).
Deterministic super-resolution regression networks typically use \(\mathcal{L}_1\) loss (Equation~\ref{eq:unet_loss}) as an objective due to its ability to produce sharper images with fewer artifacts compared to \(\mathcal{L}_2\) loss, which tends to induce blurriness and over-smoothing in the predicted output~\citep{zhaoLossFunctionsImage2017, anagunSRLibraryComparingDifferent2019, limEnhancedDeepResidual2017}.

\begin{equation}\label{eq:unet_loss}
    \mathcal{L}_1= \mathbb{E} \left[ \|y - f_\theta(x)\|_1 \right]
\end{equation}

While many different networks have been proposed for super-resolution models trained using simple deterministic regression (e.g., SRCNN~\citep{dongLearningDeepConvolutional2014}, VDSR~\citep{kimAccurateImageSuperResolution2016}, EDSR~\citep{limEnhancedDeepResidual2017}), any neural architecture that can facilitate an image-to-image mapping can be used for this task.
We adopt a U-Net architecture~\citep{ronnebergerUNetConvolutionalNetworks2015d} to serve as the core regression network for this baseline super-resolution model, and we upsample the Sentinel-2 input imagery to the desired \SIadj{2.5}{\meter} resolution using Lanczos resampling prior to passing it to the model such that the input and output images are of the same size.
The primary motivation behind this choice is that the diffusion and flow matching methods described in Sections~\ref{sec:methodology:sr:diffusion} and~\ref{sec:methodology:sr:flow_matching} use an identical U-Net architecture as the core network for their respective generative frameworks.
Consequently, using the U-Net as a straightforward regression model enables us to create a baseline where the generative benefits of these more modern approaches can be evaluated in isolation from the neural backbone's inherent capability to map low-resolution inputs to high-resolution outputs, establishing an empirical upper bound on the pixel-wise accuracy attainable.

\subsubsection{Real-ESRGAN}\label{sec:methodology:sr:esrgan}

While straightforward, simple supervised regression models are often limited in their ability to recover high-frequency details in the high-resolution image distribution that are absent from the low-resolution input~\citep{ledigPhotoRealisticSingleImage2017}.
This can be partially mitigated by integrating a perceptual loss term alongside the pixel-wise loss, where a pre-trained deep vision model \(\phi\) is used to extract high-level features from both the predicted and ground truth images, and the model is trained to minimize the distance between these feature representations in addition to the pixel-wise loss~\citep{johnsonPerceptualLossesRealTime2016}.
Perceptual feedback mechanisms introduce distortions in the pixel-wise accuracy of the predicted output, but better enable super-resolution models to produce images with sharper edges and more realistic textures that are perceptually similar to the ground truth images.
Alongside perceptual losses, adversarial learning is another effective technique for improving the perceptual quality of super-resolved images, where a ``generator'' network \(\gennetwork\) outputs synthetic high-resolution images given low-resolution inputs, and a ``discriminator'' network \(\discnetwork\) is trained to distinguish between these synthetic images and real high-resolution images~\citep{goodfellowGenerativeAdversarialNets2014, wangESRGANEnhancedSuperResolution2018}.

As a baseline adversarial super-resolution model, we used the Real-ESRGAN architecture proposed by \citet{wangRealESRGANTrainingRealWorld2021}.
The Real-ESRGAN framework utilizes three separate networks during training with the aim of maximizing both the visual quality and accuracy of generated high-resolution images:   \(\gennetwork\) learns a deterministic mapping from low-resolution inputs to high-resolution outputs, \(\discnetwork\) is trained to distinguish between real high-resolution images and those produced by \(\gennetwork\), and a fixed VGG19 network \(\phi\) pre-trained on ImageNet~\citep{imagenet_cvpr09} provides perceptual feedback to \(\gennetwork\).
\(\gennetwork\) is trained to minimize the joint loss function formulated in Equation~\ref{eq:loss_esrgan}, where \(x\) is a low-resolution input, \(y\) is a ground truth high-resolution pair, \(\phi_j\) is the \(j\)-th layer feature map of the pre-trained VGG19 network, and \(\lambda_{\text{percep}}\) and \(\lambda_{\text{adv}}\) are hyperparameters that control the relative importance of the perceptual and adversarial loss terms, respectively.

\begin{equation}\label{eq:loss_esrgan}
    \mathcal{L}_{\gennetwork} = \mathbb{E} \left[\|y - \gennetwork(x)\|_1 \right] + \lambda_{\text{percep}} \mathbb{E} \left[ \sum_j \| \phi_j(y) - \phi_j(\gennetwork(x))\|_1 \right] - \lambda_{\text{adv}} \mathbb{E} \left[ \log \discnetwork(\gennetwork(x)) \right]
\end{equation}

 Following the recommendations laid out by~\citet{wangRealESRGANTrainingRealWorld2021}, we set \(\lambda_{\text{percep}} = 1.0\) and \(\lambda_{\text{adv}} = 0.1\).
Simultaneously, \(\discnetwork\) minimizes a binary cross-entropy loss that aims to maximize the probability of correctly classifying real and generated images, as shown in Equation~\ref{eq:discriminator_loss}.

\begin{equation}\label{eq:discriminator_loss}
    \mathcal{L}_{\discnetwork} = - \mathbb{E} \left[ \log \discnetwork(y) \right]  - \mathbb{E} \left[ \log (1 - \discnetwork(\gennetwork(x))) \right]
\end{equation}

\(\gennetwork\) is a residual-in-residual dense block (RRDB) network architecture which uses convolutional layers only (i.e., no batch normalization or max-pooling layers) with many skip connections to preserve high-frequency details in the input image as it flows through the network.
 Unlike the core model used in the \(\mathcal{L}_1\) regression baseline, the RRDB network processes low-resolution images at their native resolution; upsampling does not occur until the final layers of the network, followed by two convolutional layers to produce the final synthetic high-resolution output.
\(\discnetwork\), on the other hand, uses a U-Net architecture with spectral normalization to provide pixel-level feedback on the realism of generated imagery as opposed to a single scalar output, a concept originally proposed by~\citet{schonfeldUNetBasedDiscriminator2020}.
This is a critical design choice, particularly for remote sensing imagery, as it provides the generator network with localized feedback as opposed to global feedback; given that analysis of remote sensing imagery typically concerns pixel-level information (e.g., land cover classification), this localized feedback is potentially conducive to producing higher-quality synthetic imagery.

One issue with the use of a perceptual loss term using an ImageNet pre-trained VGG19 network is that the input convolutional layer only accepts 3-band imagery, not the 4-band multispectral imagery used in this study.
 Passing only visible RGB or color infrared composites to the VGG-19 model would result in one band of the multispectral imagery being ignored (e.g., the near-infrared band in the case of RGB composites), leading to misalignment between the perceptual features extracted by the network and the actual content of the multispectral imagery.
To address this, we performed PCA on the 4-band cross-calibrated NAIP imagery in the training split of the dataset to learn a linear transformation from 4-band to 3-band imagery while preserving as much variance as possible.
Given the large size of the dataset, we used the Incremental PCA algorithm~\citep{rossIncrementalLearningRobust2008} to approximate this transformation in a memory-efficient manner by loading 1024 images at a time and updating the PCA model iteratively until all images in the training split were processed.
The first three principal components accounted for 78.55\%, 19.69\%, and 1.61\% of the variance in the 4-band NAIP imagery, respectively.
In total, the PCA transformation preserved 99.84\% of the spectral variance when mapping from 4-band to 3-band imagery.

\subsubsection{Diffusion models}\label{sec:methodology:sr:diffusion}

\paragraph{Denoising Diffusion Probabilistic Models}

While GAN-based methods became \textit{de facto} standards for super-resolution in the late 2010s, these techniques present several notable drawbacks that limit their utility.
Beyond the broader operational and practical limitations of adversarial modeling covered in Section~\ref{sec:introduction}, most variations of super-resolution GAN models only learn one-to-one mappings from a low-resolution input to a high-resolution output.
This is problematic, since a given low-resolution image can correspond to multiple potential high-resolution images~\citep{lugmayrSRFlowLearningSuperResolution2020}.
Recent research in super-resolution has increasingly shifted from adversarial and multi-loss predictive modeling frameworks toward more robust conditional generative modeling approaches~\citep{sahariaImageSuperResolutionIterative2021,liSRDiffSingleImage2022,liuDiffusionModelsMeet2024}.

Denoising Diffusion Probabilistic Models (DDPM), introduced by \citet{hoDenoisingDiffusionProbabilistic2020}, serve as the basis for modern diffusion-based generative models.
Specifically, DDPM uses a forward diffusion (noising) process that gradually adds Gaussian noise to an image \(x_0\) over \(T\) time steps to produce a sequence of increasingly noisy images \(x_1, x_2, \ldots, x_T\), where \(x_T\) is essentially pure Gaussian noise, then uses a learned reverse diffusion (i.e., denoising) process to iteratively remove noise from \(x_T\) back to \(x_0\).
Formally, the forward diffusion (i.e., noising) process \(q(x_t | x_{t-1})\) is defined as a Markov process consisting of a sequence of Gaussian transitions parameterized by a variance schedule \(\beta_1, \beta_2, \ldots, \beta_T\).
 The transition at time step (\(t\)) is given in Equation~\ref{eq:ddpm_forward}.

\begin{equation}\label{eq:ddpm_forward}
    q(x_t | x_{t-1}) = \mathcal{N}(x_t; \sqrt{1 - \beta_t} x_{t-1}, \beta_t \mathbf{I})
\end{equation}

As the variance schedule is defined \textit{a priori}, \(x_t\) can be sampled at any time step \(t\) given an initial image \(x_0\) without having to iteratively apply the forward transitions, as shown in Equation~\ref{eq:ddpm_sample_forward}, where \(\bar{\alpha}_t = \prod_{s=1}^t (1 - \beta_s)\).

\begin{equation}\label{eq:ddpm_sample_forward}
    q(x_t | x_0) = \mathcal{N}(x_t; \sqrt{\bar{\alpha}_t} x_0, (1 - \bar{\alpha}_t) \mathbf{I})
\end{equation}

Reversing the diffusion process to generate images from noise involves learning a parameterized model \(p_\theta(x_{t-1} | x_t)\) that approximates the true reverse transitions of the forward diffusion process, as shown in Equation~\ref{eq:ddpm_reverse}.

\begin{equation}\label{eq:ddpm_reverse}
    p_\theta(x_{t-1} | x_t) = \mathcal{N}(x_{t-1}; \mu_\theta(x_t, t), \beta_t \mathbf{I})
\end{equation}

During training, a random time step \(t\) can be sampled uniformly from \(\{1, \ldots, T\}\), and a noisy image \(x_t\) can be generated directly from a clean image \(x_0\) using Equation~\ref{eq:ddpm_sample_forward}.
 A neural network \(\epsilon_\theta\) is trained to predict the noise \(\epsilon\) added to \(x_0\) to produce \(x_t\) by minimizing the residual noise error.
These estimates of \(\epsilon\) can then be used to estimate the mean of the reverse transition \(\mu_\theta(x_t, t)\) as shown in Equation~\ref{eq:ddpm_model_train}.

\begin{equation}\label{eq:ddpm_model_train}
    \mu_\theta(x_t, t) = \frac{1}{\sqrt{1 - \beta_t}} \left( x_t - \frac{\beta_t}{\sqrt{1 - \bar{\alpha}_t}} \epsilon_\theta(x_t, t) \right)
\end{equation}

At inference time, starting from a sample of Gaussian noise \(x_T \sim \mathcal{N}(0, I)\), \(\mu_\theta(x_t, t)\) is used to iteratively progress backwards from \(x_T\) to \(x_0\) over \(T\) time steps by sampling from the learned reverse transitions \(p_\theta(x_{t-1} | x_t)\) (Equation~\ref{eq:ddpm_reverse}) until \(t=0\), where \(x_0\) is the generated sample.
That is to say, at each time step \(t\), the model \(\epsilon_\theta\) predicts the noise present in the current image \(x_t\) given the time step \(t\).
This noise is then scaled by a factor determined by the variance schedule and subtracted from \(x_t\) to produce image \(x_{t-1}\).
After repeating this process \(T\) times, we arrive at a generated image \(x_0\) free of noise.
While \(T\) is often fixed at 1,000 during both training and inference, \(T\) can be changed by using a strided sampling strategy during inference.
A smaller number of time steps \(T_\text{new}\) can be used by simply skipping time steps to speed up sampling at the cost of sample quality by deriving \(p_\theta(x_{t-k} | x_t)\) where \(k = T_\text{train} / T_\text{new}\) instead of \(p_\theta(x_{t-1} | x_t)\) and \(T_\text{train}\) is the number of time steps used during training.

\paragraph{Denoising Diffusion Implicit Models}

While DDPMs provide a powerful framework for generative modeling of images, they often require many inference steps to produce high-quality samples, which has hindered their adoption in practical applications.
One technique to address this limitation involves a slight modification to the reverse diffusion process, where the stochastic sampling procedure is replaced with a deterministic mapping.
Denoising Diffusion Implicit Models (DDIM), introduced by \citet{songDenoisingDiffusionImplicit2022}, essentially reframes the reverse diffusion process by removing the stochastic sampling step in the reverse transitions: instead of predicting the mean of the reverse transition to determine \(x_{t-1}\) given \(x_t\) (Equations~\ref{eq:ddpm_reverse} and~\ref{eq:ddpm_model_train}), the model's estimate of \(\epsilon_\theta(x_t, t)\) is used to derive \(x_{t-1}\) deterministically, as shown in Equation~\ref{eq:ddim_reverse}.

\begin{equation}\label{eq:ddim_reverse}
    x_{t-1} = \sqrt{\bar{\alpha}_{t-1}} \left( \frac{x_t - \sqrt{1 - \bar{\alpha}_t} \epsilon_\theta(x_t, t)}{\sqrt{\bar{\alpha}_t}} \right) + \sqrt{1 - \bar{\alpha}_{t-1}} \epsilon_\theta(x_t, t)
\end{equation}

Rather than sampling \(x_{t-1}\) from a Gaussian distribution parameterized by \(\mu_\theta(x_t, t)\) and \(\sigma_\theta(x_t, t)\), we directly compute \(x_{t-1}\) using the model's estimate of the noise present in \(x_t\).
Since no noise is added between each transition, the reverse diffusion process can be modeled as a deterministic mapping from \(x_T\) to \(x_0\).
DDIMs are designed to allow for better performance when using a reduced number of time steps during inference as the deterministic mapping allows for more consistent trajectories between \(x_T\) and \(x_0\) compared to the stochastic sampling process used in DDPMs.
As this change concerns the sampling procedure only, we can use the same trained model \(\epsilon_\theta\) from the DDPM framework without any modifications, using DDIM as a drop-in replacement for the DDPM sampling procedure during inference.

\subsubsection{Flow matching-based super-resolution}\label{sec:methodology:sr:flow_matching}

 While diffusion-based approaches are highly effective at addressing the shortcomings of prior GAN-based methods, the reverse noising process remains a complicated stochastic process that may be sub-par for actually learning robust trajectories from noise towards a Gaussian distribution.
Flow matching models, as formulated by \citet{lipmanFlowMatchingGenerative2023}, are a generalization of diffusion models that learn a deterministic forward and reverse process by modeling both as ODEs.

A continuous normalizing flow field \(v_\theta(x, t)\) is defined as the instantaneous velocity of samples as they move from an initial distribution \(x_0 \sim \mathcal{N}(0, \mathbf{I})\) to the target data distribution \(x_1 \sim p_{\text{data}}(x)\) over time \(t \in [0, 1)\), as shown in Equation~\ref{eq:flow_matching}.\footnote{When discussing flow matching, we follow the notation of \citet{lipmanFlowMatchingGenerative2023}, where \(x_1\) is a ``clean'' sample and \(x_0 \sim \mathcal{N}\), as opposed to the notation used when describing diffusion processes in Section~\ref{sec:methodology:sr:diffusion} where \(x_T \sim \mathcal{N}\) and \(x_0\) is a ``clean'' sample \citep[following][]{hoDenoisingDiffusionProbabilistic2020}.}

\begin{equation}\label{eq:flow_matching}
    \frac{dx_t}{dt} = v_\theta(x_t, t)
\end{equation}

The flow field \(v_\theta(x_t, t)\) is approximated by training a neural network to minimize \(| v_\theta(x_t, t) - \frac{d x_t}{dt} |\) over samples \(x_t\), where \(x_t\) is defined as a linear interpolation between \(x_0\) and \(x_1\) at time \(t\), as shown in Equation~\ref{eq:linear_interpolation}.

\begin{equation}\label{eq:linear_interpolation}
    x_t = (1 - t) x_0 + t x_1
\end{equation}

\citet{lipmanFlowMatchingGenerative2023} demonstrated that this linear interpolation between \(x_0\) and \(x_1\) is sufficient for image-based generative modeling tasks through the use of an image-to-image deep learning architecture for \(v_\theta\).
During training, \(x_0\) is sampled from a Gaussian distribution and \(x_1\) is sampled from the training data, and \(v_\theta\) is trained to predict the flow velocity (i.e., \(x_1 - x_0\)) at a randomly sampled time step \(t\in \left[0,1\right)\) given the interpolated sample \(x_t\) and time step \(t\) as input.
At inference time, new samples can be generated by integrating Equation~\ref{eq:flow_matching} from \(t=0\) to \(t=1\) starting from an initial sample \(x_0 \sim \mathcal{N}(0, \mathbf{I})\) and ending with a generated sample \(x_1\).
As the flow field is deterministic, any ODE solving algorithm can be used to perform this integration, and the choice of ODE solver and number of inference steps \(T\) can be chosen \textit{post hoc}.
Higher order solvers use multiple neural function evaluations (i.e., forward passes) of the flow field \(v_\theta(x_t, t)\) at each time step to better approximate the integral of Equation~\ref{eq:flow_matching} at each step, resulting in more accurate sample trajectories at the cost of increased computation compared to lower-order methods.
A summary of the ODE solvers used in this study is provided in Table~\ref{tab:ode_solvers}, where \(\Delta t = \frac{1}{T}\).

\begin{table}[H]
    \centering
    \caption{Summary of ODE solvers used for flow matching model inference.}\label{tab:ode_solvers}
    \begin{tabular}{l l c}
        \toprule
        \textbf{Method} & \textbf{Update Equation} & \textbf{Order} \\
        \midrule
        Euler & \(x_{t+\Delta t} = x_t + v_\theta(x_t, t) \Delta t\) & 1 \\
        \addlinespace
        Midpoint & \(x_{t+\Delta t} = x_t + v_\theta\left(x_t + \frac{1}{2} v_\theta(x_t, t) \Delta t, t + \frac{1}{2} \Delta t\right) \Delta t\) & 2 \\
        \addlinespace
        Heun & \(x_{t+\Delta t} = x_t + \frac{1}{2} \left( v_\theta(x_t, t) + v_\theta\left(x_t + v_\theta(x_t, t) \Delta t, t + \Delta t\right) \right) \Delta t\) & 2 \\
        \addlinespace
        RK4 &
        \begin{tabular}{l}
            \(k_1 = v_\theta(x_t, t)\) \\
            \(k_2 = v_\theta\left(x_t + \frac{1}{2} k_1 \Delta t, t + \frac{1}{2} \Delta t\right)\) \\
            \(k_3 = v_\theta\left(x_t + \frac{1}{2} k_2 \Delta t, t + \frac{1}{2} \Delta t\right)\) \\
            \(k_4 = v_\theta\left(x_t + k_3 \Delta t, t + \Delta t\right)\) \\
            \(x_{t+\Delta t} = x_t + \frac{1}{6} (k_1 + 2k_2 + 2k_3 + k_4) \Delta t\)
        \end{tabular} & 4 \\
        \bottomrule
    \end{tabular}
\end{table}

\subsubsection{ Network architecture, conditioning, and model training}\label{sec:methodology:sr:training}

The flow matching and diffusion networks (\(v_\theta\) and \(\epsilon_\theta\)) used the same U-Net backbone as the supervised \(\mathcal{L}_1\) regression baseline, with sinusoidal timestep embeddings followed by a learned projection to provide temporal conditioning~\citep{hoDenoisingDiffusionProbabilistic2020,vaswaniAttentionAllYou2017}.
We opted to use \(\mathcal{L}_1\) loss instead of the standard \(\mathcal{L}_2\) loss during training, as the ablation study performed by \citet{sahariaImageSuperResolutionIterative2021} showed that training with \(\mathcal{L}_1\) loss produced sharper and more perceptually realistic super-resolved images than \(\mathcal{L}_2\) loss (similar to the reasoning behind using \(\mathcal{L}_1\) loss for the supervised regression baseline, as discussed in Section~\ref{sec:methodology:sr:unet_baseline}).
Pursuant to this, the diffusion network \(\epsilon_\theta\) was trained to predict the injected noise by minimizing \(\mathcal{L}_{\epsilon_\theta} = \mathbb{E} \left[ \|\epsilon - \epsilon_\theta(x_t, t)\|_1 \right]\), where \(\epsilon\) is the Gaussian noise added to the clean image \(x_0\) to produce the noisy image \(x_t\).
Likewise, the flow matching velocity network \(v_\theta\) was trained to minimize the residual flow error by minimizing \(\mathcal{L}_{v_\theta} = \mathbb{E} \left[ \|v_\theta(x_t, t) - (x_1 - x_0)\|_1 \right]\), where \(x_1 - x_0\) is the true flow velocity between the initial and target distributions at time step \(t\).

\begin{figure}[H]
    \centering
    \includegraphics[width=\textwidth]{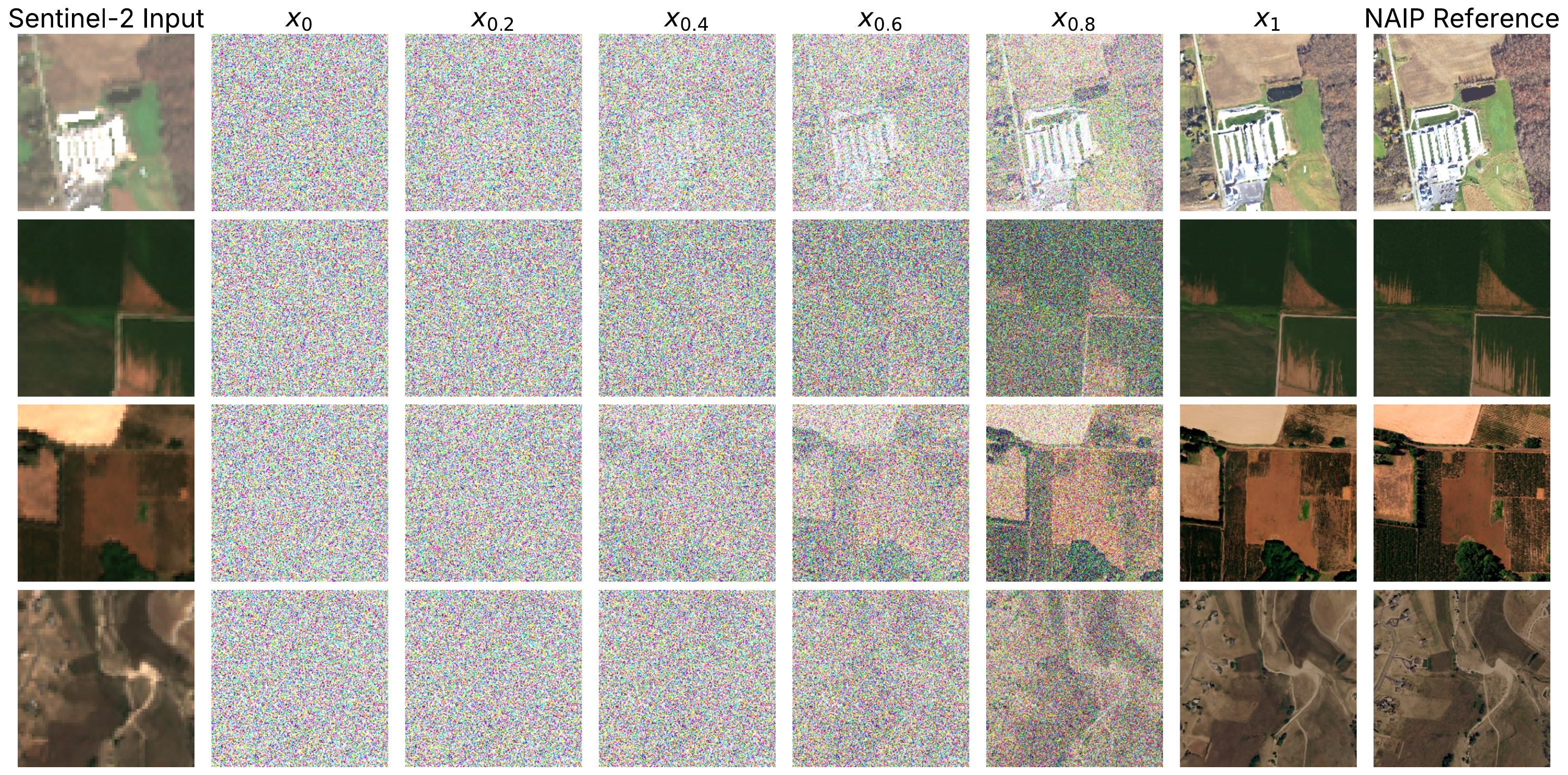}
    \caption{Visualization of the sampling procedure for flow matching super-resolution using the Euler solver. Starting from pure Gaussian noise (\(x_0\)), the model progressively refines the sample over \(T\) steps until reaching the final super-resolved output (\(x_1\)).}\label{fig:fm_sampling}
\end{figure}

Extending the unconditional generative modeling mechanisms of the diffusion and flow matching models to super-resolution requires incorporating a conditioning mechanism to guide the networks' generation.
Under the SR3 framework, \citet{sahariaImageSuperResolutionIterative2021} showed that simply concatenating the low-resolution input image to the noisy image along the channel dimension at each time step was an effective conditioning mechanism for super-resolution tasks.
To allow channel-wise concatenation, the low-resolution Sentinel-2 input was upsampled to match the \SIadj{2.5}{\meter} target grid using Lanczos resampling.
Figure~\ref{fig:fm_sampling} illustrates the effect of the conditioning mechanism for the flow matching model, demonstrating how the upsampled Sentinel-2 imagery guides the generation process from the source distribution (Gaussian noise) to the target distribution (\SIadj{2.5}{\meter} NAIP imagery).

All super-resolution models (flow matching, diffusion, Real-ESRGAN, and the supervised \(\mathcal{L}_1\) regression baseline) were trained using an identical optimization protocol for 100 epochs.
We used the AdamW optimizer with an initial learning rate of \(1 \times 10^{-5}\) that linearly increased to \(1 \times 10^{-4}\) over the first 10 epochs, followed by a cosine annealing schedule decreasing the learning rate to 0 over the remaining 90 epochs.
Training batches were set to 16 images per forward pass, with gradients accumulated across 2 batches prior to weight updates to yield an effective batch size of 32.
Automatic mixed precision with \texttt{bfloat16} half-precision was implemented to minimize VRAM footprint and accelerate throughput.

\subsection{Evaluation of super-resolution models}\label{sec:methodology:sr_evaluation}

Once trained, we evaluated the performance of each super-resolution model on the validation split of the Sentinel-2/NAIP pair dataset.
For the diffusion and flow matching models, we specifically tested various step sizes to assess the trade-off between inference speed and super-resolution quality.
 We tested \(T \in \{1, 5, 10, 15, \dots, 100\}\) time steps for both diffusion (DDPM and DDIM) and flow matching models (with all ODE solvers) to evaluate this trade-off, whereas Real-ESRGAN and the supervised \(\mathcal{L}_1\) U-Net baseline require only a single forward pass.
For broad assessment of pixel-wise accuracy, we used the Peak Signal-to-Noise Ratio (PSNR) metric, defined in Equation~\ref{eq:psnr}, where \(L\) is the range of possible pixel values (\(L=2\) for our as our images are normalized to \([-1, 1]\)), \(y\) is the ground truth image, \(\hat{y}\) is the super-resolved image, and \(N\) is the total number of pixels in the image.

\begin{equation}\label{eq:psnr}
    \text{PSNR} = 10 \log_{10} \left( \frac{L^2}{{||y - \hat{y}||_2 / N}} \right)
\end{equation}

PSNR is useful for measuring overall pixel-wise accuracy in a way that is easy to interpret due to its logarithmic scale; for example, a 1 dB increase in PSNR corresponds to a 10\% decrease in MSE.
Yet, PSNR correlates poorly with human perception when used as a measure of image similarity for super-resolution tasks~\citep{ledigPhotoRealisticSingleImage2017}.
Producing visually realistic super-resolved imagery at 4\(\times\) the original resolution forces the model to hallucinate fine details absent in the low-resolution input.
While these details are ``correct'' from a visual standpoint and are often a desired outcome of super-resolution, they tend to increase pixel-wise error compared to the ground truth high-resolution image, thus lowering PSNR~\citep{blauPerceptionDistortionTradeoff2018}.
In this context, the structural similarity index measure (SSIM) and learned perceptual image patch similarity (LPIPS) are designed to quantify similarity in image structure and visual perception as opposed to pixel-wise accuracy.
SSIM, as formulated in Equation~\ref{eq:ssim}, uses a local Gaussian kernel (typically 11\(\times\)11 pixels with \(\sigma=1.5\)) to compute local statistics (mean \(\mu\), variance \(\sigma^2\), and covariance \(\sigma_{xy}\)) between two images \(x\) and \(y\), with constants \(k_1 = 0.01\) and \(k_2 = 0.03\) used for numerical stability~\citep{wangImageQualityAssessment2004}.

\begin{equation}\label{eq:ssim}
    \text{SSIM}(x, y) = \frac{(2 \mu_x \mu_y + (k_1L)^2)(2 \sigma_{xy} + (k_2L)^2)}{(\mu_x^2 + \mu_y^2 + (k_1L)^2)(\sigma_x^2 + \sigma_y^2 + (k_2L)^2)}
\end{equation}

The Gaussian kernel is slid across the entire image (with 1 pixel stride) to compute a mean SSIM value for the entire image.
Values for SSIM range from -1 to 1, with values closer to 1 indicating greater structural similarity between the two images.
LPIPS uses an ImageNet pre-trained network to extract deep feature maps from multiple layers of the network for both images \(x\) and \(y\).
As the feature maps pass through the network, the \(\mathcal{L}_2\) distance between corresponding feature maps is computed after selected layers \(l\).
Equation~\ref{eq:lpips} formulates the LPIPS metric, where \(H_l\) and \(W_l\) are the height and width of the feature maps at layer \(l\), respectively, and \(w_l\) is a learned weight for layer \(l\) that adjusts the importance of each layer's contribution to the final LPIPS score.

\begin{equation}\label{eq:lpips}
    \text{LPIPS}(x, y) = \sum_l \frac{w_l}{H_l W_l} \sum_{h=1}^{H_l} \sum_{w=1}^{W_l} \|\phi_l(x)_{h,w} - \phi_l(y)_{h,w}\|_2^2
\end{equation}

We used an AlexNet-based implementation of LPIPS where weights \(w_l\) are trained using a human-annotated dataset of image similarity judgments to better align the feature relevance of each layer with human perception~\citep{krizhevskyImageNetClassificationDeep2012}.
As the AlexNet architecture only accepts 3-band imagery, we used the same PCA transformation described previously to reduce the 4-band calibrated NAIP imagery to 3-band imagery prior to LPIPS calculation.
LPIPS was used to determine the optimal value of \(T\) for diffusion and flow matching models using the Kneedle algorithm~\citep{satopaaFindingKneedleHaystack2011} to identify the elbow point where increasing \(T\) yielded diminishing returns in LPIPS improvement, providing an objective method for selecting \(T\) that balances inference speed and perceptual quality.
Additionally, once we identified the best performing super-resolution model according to PSNR (separate from the LPIPS-based selection of \(T\)), we performed a spectral evaluation of the super-resolved imagery using the metrics in Table~\ref{tab:spectral_metrics} to quantify the agreement between the super-resolved imagery and the cross-calibrated NAIP imagery on a per-band basis to assess the spectral reliability of the super-resolved imagery.

\begin{table}[H]
    \centering
    \caption{Spectral similarity metrics used to assess the reliability of super-resolved imagery.}\label{tab:spectral_metrics}
    \begin{tabular}{l l}
        \toprule
        \textbf{Metric} & \textbf{Formula} \\
        \midrule
        \(R^2\) & \(1 - \frac{\sum_{i=1}^N (y_i - \hat{y}_i)^2}{\sum_{i=1}^N (y_i - \bar{y})^2}\) \\
        \addlinespace
        Root Mean Squared Error (RMSE) & \(\sqrt{\frac{1}{N} \sum_{i=1}^N (y_i - \hat{y}_i)^2}\) \\
        \addlinespace
        Mean Absolute Error (MAE) & \(\frac{1}{N} \sum_{i=1}^N |y_i - \hat{y}_i|\) \\
        \addlinespace
        Mean Absolute Percentage Error (MAPE) & \(\frac{100}{N} \sum_{i=1}^N \left| \frac{y_i - \hat{y}_i}{y_i} \right|\) \\
        \bottomrule
    \end{tabular}
\end{table}

\subsection{CONUS Sentinel-2 super-resolution product}\label{sec:methodology:conus_sr}

Upon identifying the best performing super-resolution model and inference configuration based on the pixel-wise metrics (as opposed to the perceptual metrics), we used this model to generate super-resolved Sentinel-2 imagery for the entire CONUS region.
We disregarded the perceptual metrics for this process: a synthetic imagery product should be spectrally reliable in order to be suitable for additional analysis, and while high perceptual quality is desirable, we prioritized spectral reliability for the creation of our super-resolved Sentinel-2 product.
We used the Sentinel-2 Level-2A annual composites generated in Section~\ref{sec:methodology:preprocessing:composites} as inputs to the super-resolution model.

As each Sentinel-2 tile is approximately 10,980\(\times\)10,980 pixels at \SIadj{10}{\meter} resolution (or 43,920\(\times\)43,920 pixels at \SIadj{2.5}{\meter} resolution), we utilized a sliding window approach for inference as opposed to super-resolving the entire tile in a single forward pass.
We used a window size of 64\(\times\)64 pixels at \SIadj{10}{\meter} resolution (or 256\(\times\)256 pixels at \SIadj{2.5}{\meter} resolution) with a stride of half the window size to ensure that each pixel is covered by multiple windows and to mitigate edge artifacts in the super-resolved output.
The super-resolved outputs were weighted according to a 2D Gaussian kernel centered at the middle of the window that decreases the influence of edge pixels when combining the super-resolved outputs from overlapping windows.
We performed this procedure for each of the 988 Sentinel-2 tiles across CONUS to produce a complete super-resolved Sentinel-2 product for the year 2025, ultimately comprising 1.58 trillion pixels of synthetic imagery.

\subsection{Demonstrating utility for land cover classification}\label{sec:methodology:lc}

To demonstrate the practical utility of our super-resolution model for downstream applications, we used the model outputs to train classifiers for land cover mapping at \SIadj{2.5}{\meter} resolution.
Using the Sentinel-2/NAIP/CBLC triplet dataset curated in Section~\ref{sec:methodology:preprocessing:cbp}, we compared the performance of various semantic segmentation models trained to predict land cover labels with the super-resolved Sentinel-2 imagery as inputs.
 To determine the best performing super-resolution framework (e.g., sampling process, \(T\) value, ODE solver, etc.) for land cover classification, we generated super-resolved Sentinel-2 imagery using all super-resolution models and configurations with \(T \in \{1, 10, 20, 30, 40, 50\}\) for all samples in the training and test splits of the Sentinel-2/NAIP/CBLC dataset (except for Real-ESRGAN and the supervised \(\mathcal{L}_1\) U-Net baseline, which do not have a \(T\) parameter).
We used 3 different semantic segmentation architectures to ensure that our findings were not architecture-specific: U-Net, DeepLabV3+~\citep{chenEncoderDecoderAtrousSeparable2018}, and SegFormer~\citep{xieSegFormerSimpleEfficient2021}.
The U-Net and DeepLabV3+ models utilized a ResNet-101 CNN encoder~\citep{heDeepResidualLearning2015}, while the SegFormer model used an MiT-B5 Transformer encoder~\citep{xieSegFormerSimpleEfficient2021}.
The encoder of each architecture was initialized with ImageNet pre-trained weights, while the decoder was randomly initialized.
Focal loss was used to address the class imbalance present in the CBLC dataset, as shown in Equation~\ref{eq:focal_loss}, where \(y_i\) is the ground truth one-hot encoded label for pixel \(i\), \(\hat{y}_i\) is the predicted probability for the true class, and \(\gamma\) is a hyperparameter that determines the strength of the modulating factor (we used \(\gamma = 2\) in this study, per the original Focal loss implementation)~\citep{linFocalLossDense2020}.

\begin{equation}\label{eq:focal_loss}
    \mathcal{L}_{\text{focal}} = - \sum_{i=1}^{N} (1 - \hat{y}_i)^\gamma y_i \log(\hat{y}_i)
\end{equation}

 Downstream segmentation models were trained using the same optimization protocol described in Section~\ref{sec:methodology:sr:training}: AdamW optimizer with a 10-epoch linear warmup to \(1 \times 10^{-4}\), cosine decay to 0 over the remaining 90 epochs (100 total epochs), batch size 16 with gradient accumulation across 2 batches (effective batch size 32), and \texttt{bfloat16} automatic mixed precision.
5-fold cross-validation was used on the training split; loss was calculated on the validation fold after each epoch, and the model obtained in the epoch with the lowest validation loss was used to evaluate performance on the test split.
We used the classification metrics summarized in Table~\ref{tab:classification_metrics} to evaluate land cover classification performance on the test split, where \(\text{TP}\), \(\text{FP}\), and \(\text{FN}\) are the number of true positive, false positive, and false negative predictions, respectively.

\begin{table}[H]
    \centering
    \caption{Classification metrics used to evaluate land cover classification performance.}\label{tab:classification_metrics}
    \begin{tabular}{l l}
        \toprule
        \textbf{Metric} & \textbf{Formula} \\
        \midrule
        User's Accuracy (UA) & \(\text{UA} = \frac{\text{TP}}{\text{TP} + \text{FP}}\) \\
        \addlinespace
        Producer's Accuracy (PA) & \(\text{PA} = \frac{\text{TP}}{\text{TP} + \text{FN}} \) \\
        \addlinespace
        F1 Score (F1) & \(\text{F1} = \frac{2 \times \text{TP}}{2 \times \text{TP} + \text{FP} + \text{FN}}\) \\
        \addlinespace
        Overall Accuracy (OA)  &  \(\text{OA} = \frac{\sum \text{TP}}{N_\text{total}}\) \\
        \bottomrule
    \end{tabular}
\end{table}

When reporting metrics on the test split, we averaged the metrics across all 5 cross-validation runs to obtain a robust estimate of the true performance of each test case.
 For class-averaged metrics (user's/producer's accuracy, F1 Score), each class contributes equally to the macro-average regardless of the number of pixels belonging to each class, whereas overall accuracy is computed globally across all pixels without regard for class.
Once the best performing super-resolution model and configuration were identified for land cover classification according to the  mean class-averaged F1 score across the folds, we compared its classification performance against models trained using the original low-resolution Sentinel-2 imagery (upsampled to \SIadj{2.5}{\meter} using Lanczos resampling) and the original high-resolution NAIP imagery.

We used the best-performing super-resolution model, inference parameters, and semantic segmentation model to generate annual land cover maps at \SIadj{2.5}{\meter} resolution for the entire Chesapeake Bay watershed for 2020--2025, using 6 years of annual mean Sentinel-2 temporal composites as input to the super-resolution model (Section~\ref{sec:methodology:preprocessing:composites}), which then directly feeds into the trained land cover classification model.
This process was similar to the CONUS-scale synthetic imagery generation process described in Section~\ref{sec:methodology:conus_sr}, where we used a sliding window approach with a window size of 64\(\times\)64 pixels at \SIadj{10}{\meter} resolution and a stride of 32 pixels to ensure that each pixel was covered by multiple windows.
The \SIadj{10}{\meter} data within each window was super-resolved to \SIadj{2.5}{\meter} resolution using the chosen super-resolution technique, then passed through the trained land cover classification model to produce a land cover probability map for each window.
Like the super-resolution process, the land cover probability maps for each window were weighted according to a 2D Gaussian kernel centered at the middle of the window to smooth the predictions across window edges.
Finally, each pixel's land cover label is determined by the class with the highest predicted probability for that pixel (post-weighting).
We process and store these land cover maps in raster format according to the original Sentinel-2 tiling scheme, but clip the output rasters to the boundary of the Chesapeake Bay watershed: ultimately, each year of land cover product consists of 50 Sentinel-2 tiles covering the entire Chesapeake Bay watershed at \SIadj{2.5}{\meter} resolution, which amounts to over 357 billion pixels across 6 years of data.

Once the land cover maps were generated, we performed a point-based accuracy assessment using the locations and labels derived from the CBLC dataset as reference data (Section~\ref{sec:methodology:preprocessing:cbp}).
As our product is multi-temporal while the assessment points only provide labels for 2021 and 2022, we used the appropriate year's land cover map for each assessment point when determining the predicted land cover label for that location , bounding our assessment to 2021 and 2022 data only.
In addition to the standard classification metrics described in Table~\ref{tab:classification_metrics}, we also calculated the confusion matrix for each class to provide a more detailed breakdown of the types of errors made by the model.

\section{Results}

\subsection{Evaluation of super-resolution performance}

\begin{figure}[H]
    \centering
    \includegraphics[width=\textwidth]{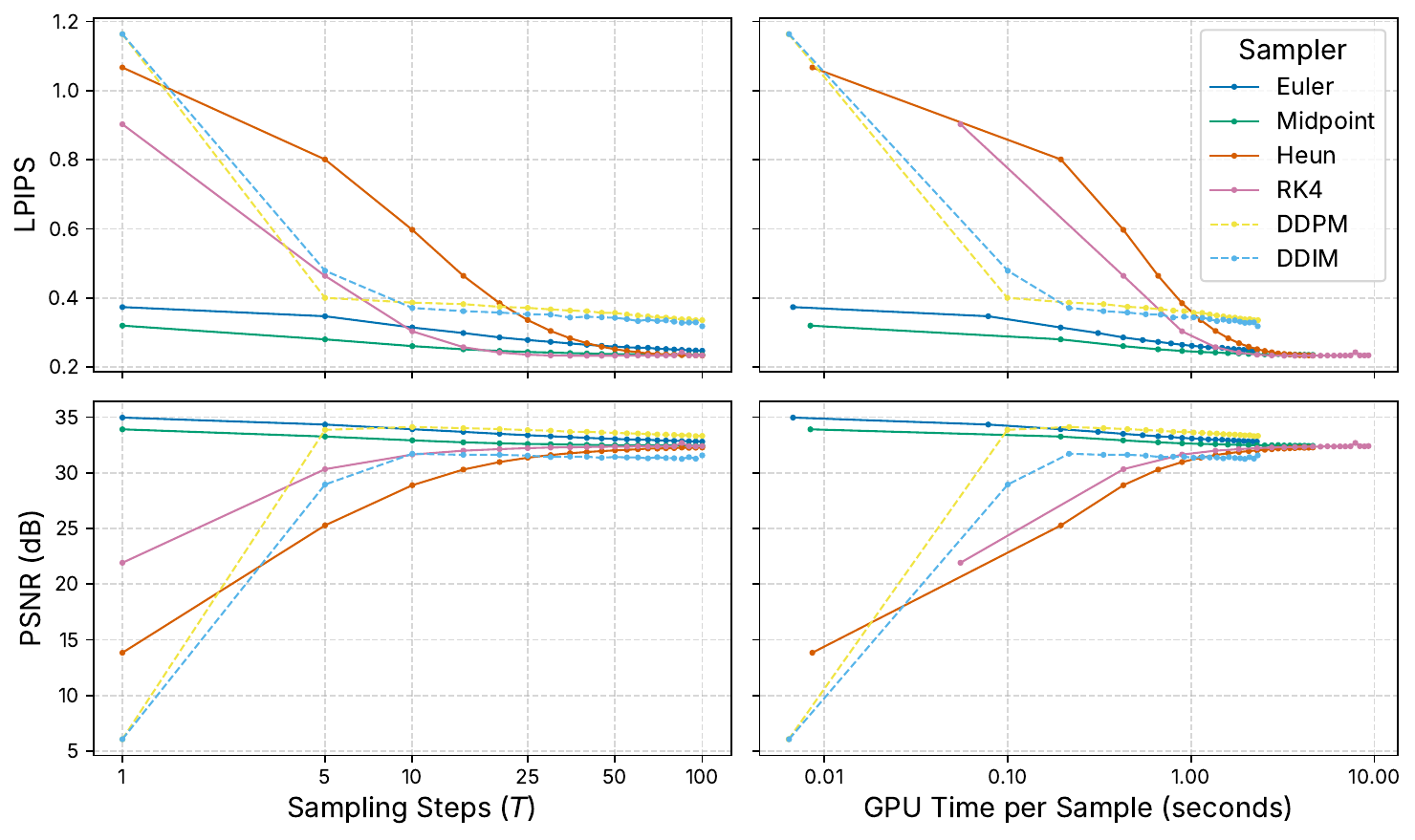}
    \caption{Pixel-wise accuracy (PSNR) and perceptual similarity (LPIPS) performance of the diffusion and flow matching-based super-resolution models across different ODE solvers/diffusion samplers and numbers of sampling steps \(T\).}\label{fig:sampling_performance}
\end{figure}

First, we evaluated the performance of the diffusion and flow matching super-resolution models across different sampling algorithms and numbers of sampling steps \(T\) on the validation set.
Figure~\ref{fig:sampling_performance} shows the results of the flow matching ODE solvers (Table~\ref{tab:ode_solvers}) and DDIM/DDPM samplers across different values of \(T\) in terms of both pixel-wise accuracy (PSNR) and perceptual quality (LPIPS) metrics, as well as their inference time.
When examining the outputs of the diffusion-based samplers, the DDIM sampler was consistently outperformed by the DDPM sampler at all values of \(T\) when considering PSNR.
In terms of LPIPS, DDIM slightly outperformed the DDPM sampler where \(T \geq 10\), with the DDPM sampler ultimately achieving LPIPS and PSNR values of 0.3358 and 33.31 dB, respectively, at \(T=100\), compared to 0.3184 and 31.56 dB for the DDIM sampler.
However, the diffusion-based samplers were unable to match the performance of the flow matching-based models across any of the ODE solvers tested, regardless of the number of sampling steps used.
Additionally, both diffusion-based samplers failed to produce any meaningful results at \(T=1\), both yielding PSNR values of 6.07 dB and LPIPS of 1.1637 (DDPM and DDIM are equivalent at \(T=1\)).

Overall, the Euler and Midpoint ODE solvers substantially outperformed all other ODE solvers and both diffusion samplers at \(T \leq 5\), with the Midpoint producing slightly better perceptual performance at the cost of pixel-wise accuracy, with average PSNR and LPIPS decreases of 1.067 dB and 0.0603, respectively, compared to the Euler method.
The RK4 solver outperformed the Euler solver in terms of LPIPS at \(T=10\) and outperformed all other solvers at \(T=20\), but never surpassed the Euler or Midpoint solver in terms of PSNR except for \(T=85\), where it surpassed the Midpoint solver before underperforming at higher values of \(T\).
Indeed, the RK4 achieved the best LPIPS at \(T=40\) with a value of 0.2330, but at the cost of inference time, requiring an average of 3.69 seconds per sample.
The benefits of using the more basic Euler and Midpoint solvers were clear when comparing the GPU time per sample during inference, as the Heun and RK4 solvers did not approach comparable performance with these simpler methods until the time needed to sample a single image exceeded 1 second.
In fact, the Euler and Midpoint solvers achieved relatively high performance at \(T=1\), with the Midpoint solver achieving an LPIPS of 0.3200 and PSNR of 33.90 dB, whereas the Euler solver produced a PSNR of 34.95 dB but a relatively poor LPIPS of 0.3738.
The ability to generate high-quality, pixel-accurate super-resolved imagery in a single step was a key advantage of the flow matching framework over diffusion-based models, as high spectral reliability and low inference time are conducive to real-world remote sensing applications.

\begin{figure}[H]
    \centering
    \includegraphics[width=\textwidth]{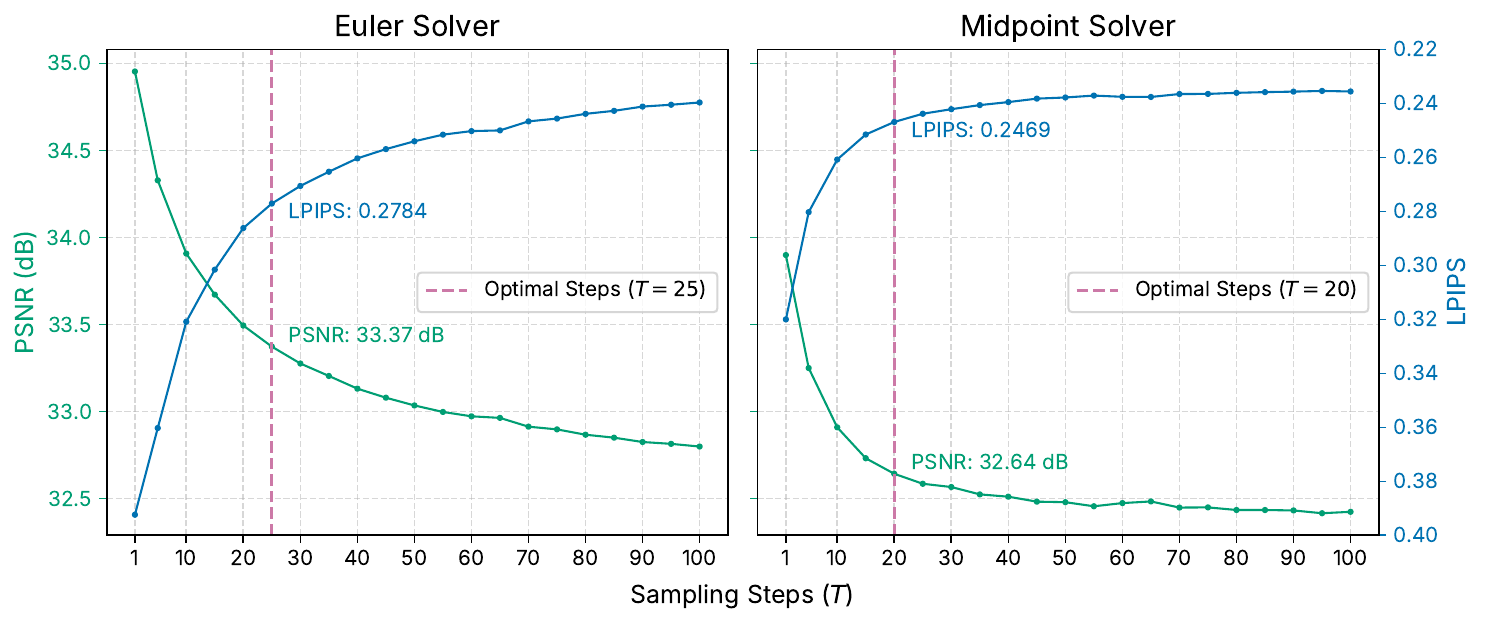}
    \caption{Comparison of Euler and Midpoint solvers for flow matching-based super-resolution in terms of pixel-wise accuracy (PSNR) and perceptual quality (LPIPS) across varying numbers of sampling steps \(T\). Note that the y-axis for LPIPS is inverted for easier visualization.}\label{fig:euler_vs_midpoint}
\end{figure}

\begin{figure}[H]
    \centering
    \includegraphics[width=\textwidth]{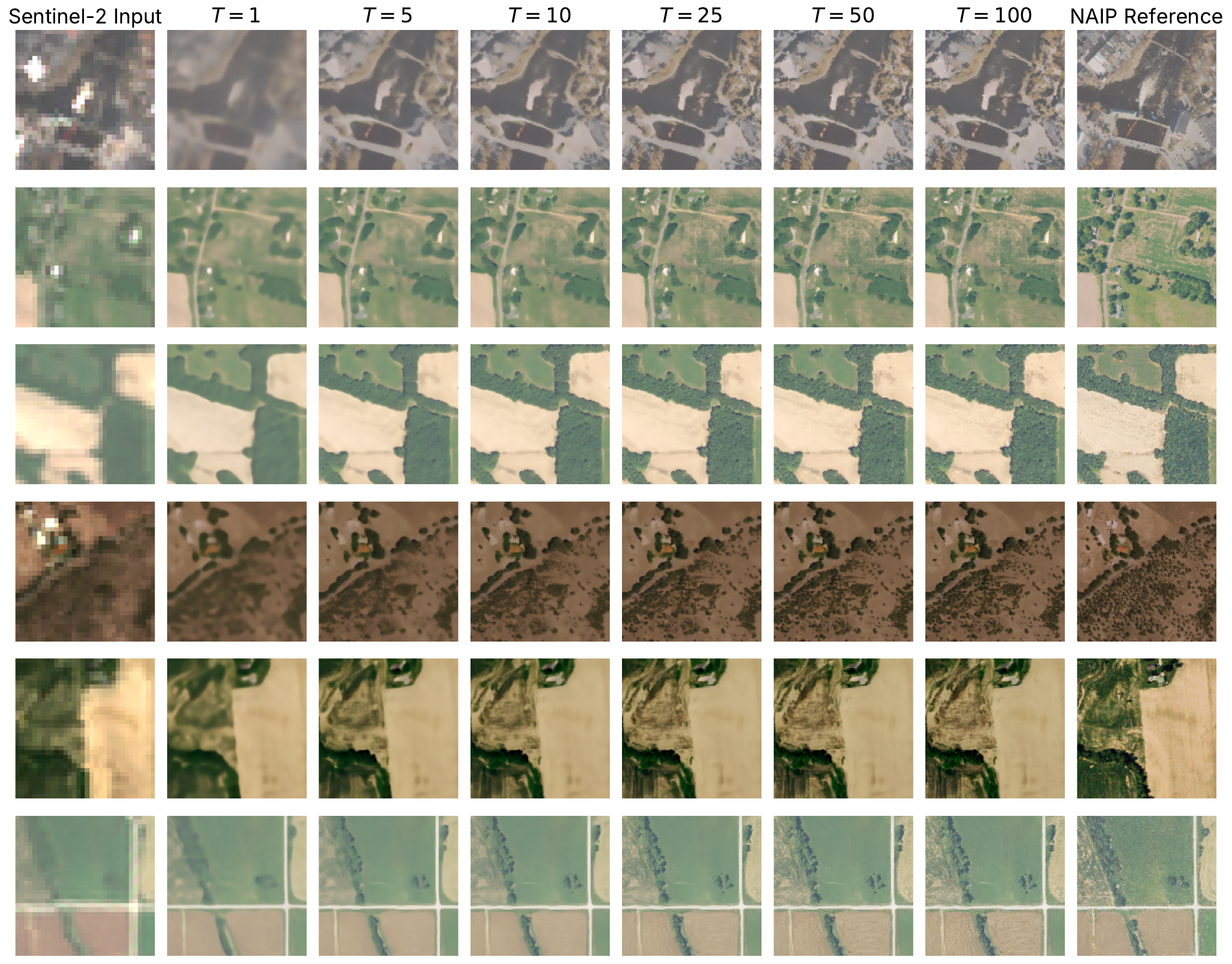}
    \caption{
        Visual comparison of super-resolved outputs from the flow matching model using the Euler solver with varying numbers of sampling steps \(T\). As \(T\) increases, more fine details are synthesized in the super-resolved output images at the cost of pixel-wise accuracy.
    }\label{fig:fm_var_steps}
\end{figure}

To examine the differences between the Euler and Midpoint solvers more closely, we provide a more detailed comparison of their performance across different values of \(T\) in Figure~\ref{fig:euler_vs_midpoint}.
The first-order Euler method yielded much higher PSNR values than the second-order Midpoint method while the Midpoint solver was superior in terms of LPIPS when comparing the two solvers at equal \(T\).
At the optimal value of \(T\) for both solvers (found using the Kneedle algorithm), the Midpoint method achieved a mean LPIPS of 0.2469 compared to 0.2784 for the Euler method, while the Euler method yielded a mean PSNR of 33.37 dB compared to 32.64 dB for the Midpoint method.
Perhaps most notable was the inverse relationship between PSNR and LPIPS: as the number of sampling steps increased, the pixel-wise accuracy of the outputs decreased while the perceptual quality increased.
We used the Euler sampling method to illustrate this trade-off in Figure~\ref{fig:fm_var_steps}: at \(T=1\), the output image exhibits sharper edges between different objects and land cover types, yet lacks finer details (e.g., the texture of trees in a forest canopy).
As \(T\) increased, more fine details were synthesized to produce a visually pleasing image, but this came at the cost of spectral accuracy as the model must synthesize these details according to patterns learned during training as opposed to directly reconstructing them from the low-resolution input image.
Being able to directly model the perception-distortion trade-off and have fine control over the quality of the synthesized images at inference time is an advantage of the flow matching framework compared to the diffusion models or the GAN-based Real-ESRGAN model.

\begin{figure}[H]
    \centering
    \includegraphics[width=\textwidth]{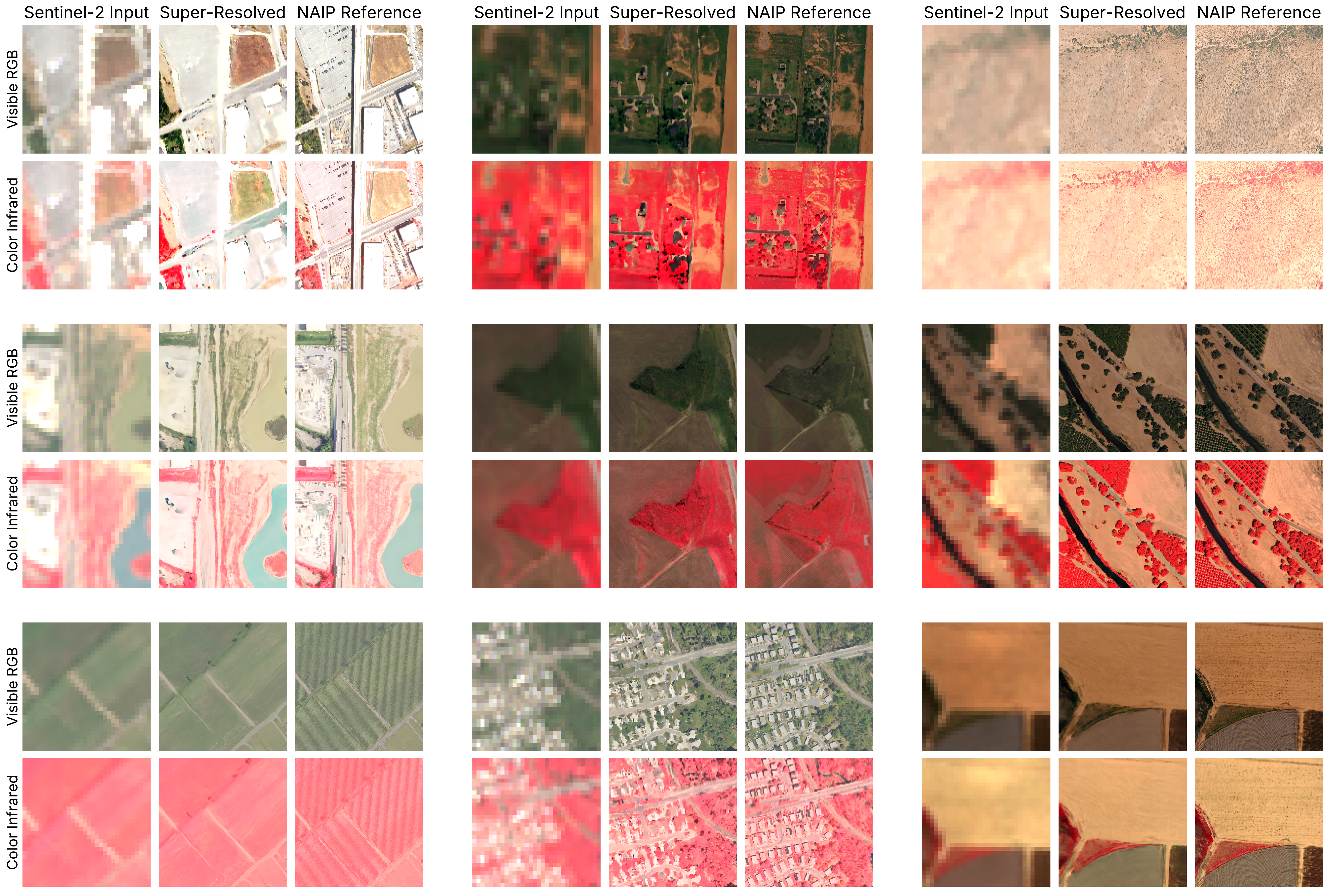}
    \caption{\SIadj{10}{\meter} Sentinel-2 imagery, synthetic \SIadj{2.5}{\meter} super-resolved imagery produced by the flow matching model using the Midpoint solver with \(T=20\), and true \SIadj{2.5}{\meter} NAIP imagery for comparison.}\label{fig:sr_outputs}
\end{figure}

Table~\ref{tab:sr_results} provides the mean PSNR, SSIM, and LPIPS values across the validation set for all super-resolution models, samplers, and steps \(T=1\), \(T=100\), and the optimal \(T\) as determined by the Kneedle algorithm.
 The supervised \(\mathcal{L}_1\) regression U-Net baseline established the upper bound for pixel-level fidelity among all configurations, yielding a mean PSNR of 35.32 dB and an SSIM of 0.8377.
However, this came at the expense of perceptual realism, with the U-Net regression baseline posting the highest LPIPS among deep learning approaches (0.4437).
The flow matching model sampled with the single-step Euler solver (\(T=1\)) closely approached the regression baseline, achieving a mean PSNR of 34.95 dB and SSIM of 0.8316 (within 0.37 dB and 0.0061 of the baseline, respectively) while yielding an improved LPIPS of 0.3738.
Both methods outperformed Real-ESRGAN (33.29 dB PSNR, 0.7861 SSIM) and Lanczos upsampling (32.99 dB PSNR, 0.8134 SSIM) in pixel-wise metrics.
The Midpoint solver at its optimal \(T=20\) appeared to have the best balance between LPIPS and inference time among the iterative refinement methods, achieving a mean LPIPS of 0.2469 and an SSIM of 0.7714 at an average inference time of 0.894 GPU seconds per sample; we provide some examples of the outputs from this model in Figure~\ref{fig:sr_outputs}.
 Among the model-driven methods, the Real-ESRGAN model was the fastest in terms of inference time at 0.00274 GPU seconds per sample.

\begin{table}[H]
    \centering
    \caption{Spectral and visual quality metrics for super-resolution results across various models, samplers, and numbers of sampling steps \(T\).}\label{tab:sr_results}
    \begin{tabular*}{\textwidth}
    {@{\extracolsep{\fill}}
        l c S[table-format=2.2] S[table-format=1.4] S[table-format=1.4] r @{}}
        \toprule
        Method/Sampler & Steps ($T$) & {PSNR (dB) $\uparrow$} & {SSIM $\uparrow$} & {LPIPS $\downarrow$} & \multicolumn{1}{c}{GPU Time/Sample (s) $\downarrow$} \\
        \midrule
        Lanczos & -- & 32.99 & 0.8134 & 0.9065 & -- \\
        \addlinespace

        \(\mathcal{L}_1\) Regression U-Net
        & --
        & \multicolumn{1}{c}{35.32}
        & \multicolumn{1}{c}{0.8377}
        & \multicolumn{1}{c}{0.4437}
        & 6.24 $\times 10^{-3}$
        \\

        Real-ESRGAN & -- & 33.29 & 0.7861 & 0.1937 & 2.74 $\times 10^{-3}$ \\

        \addlinespace
        \multirow{3}{*}{DDPM} & 1 & 6.07 & 0.0000 & 1.1637 & 6.39 $\times 10^{-3}$ \\
            & 5 & 33.85 & 0.8232 & 0.4004 & 9.96 $\times 10^{-2}$ \\
            & 100 & 33.31 & 0.8027 & 0.3358 & 2.32 $\times 10^{+0}$ \\
        \addlinespace
        \multirow{3}{*}{DDIM} & 1 & 6.07 & 0.0000 & 1.1637 & 6.42 $\times 10^{-3}$ \\
            & 10 & 31.70 & 0.7338 & 0.3713 & 2.16 $\times 10^{-1}$ \\
            & 100 & 31.56 & 0.7366 & 0.3184 & 2.31 $\times 10^{+0}$ \\
        \addlinespace
        \multirow{3}{*}{Euler} & 1 & 34.95 & 0.8316 & 0.3738 & 6.75 $\times 10^{-3}$ \\
            & 25 & 33.37 & 0.7958 & 0.2784 & 5.45 $\times 10^{-1}$ \\
            & 100 & 32.80 & 0.7744 & 0.2474 & 2.29 $\times 10^{+0}$ \\
        \addlinespace
        \multirow{3}{*}{Midpoint} & 1 & 33.90 & 0.8178 & 0.3200 & 8.40 $\times 10^{-3}$ \\
            & 20 & 32.64 & 0.7714 & 0.2469 & 8.94 $\times 10^{-1}$ \\
            & 100 & 32.42 & 0.7603 & 0.2356 & 4.62 $\times 10^{+0}$ \\
        \addlinespace
        \multirow{3}{*}{Heun} & 1 & 13.84 & 0.0281 & 1.0672 & 8.60 $\times 10^{-3}$ \\
            & 25 & 31.35 & 0.7046 & 0.3362 & 1.13 $\times 10^{+0}$ \\
            & 100 & 32.26 & 0.7557 & 0.2336 & 4.62 $\times 10^{+0}$ \\
        \addlinespace
        \multirow{3}{*}{RK4} & 1 & 21.91 & 0.1519 & 0.9030 & 5.53 $\times 10^{-2}$ \\
            & 15 & 31.98 & 0.7438 & 0.2577 & 1.36 $\times 10^{+0}$ \\
            & 100 & 32.40 & 0.7588 & 0.2342 & 9.29 $\times 10^{+0}$ \\
        \bottomrule
    \end{tabular*}
\end{table}

\begin{figure}[H]
    \centering
    \includegraphics[width=\textwidth]{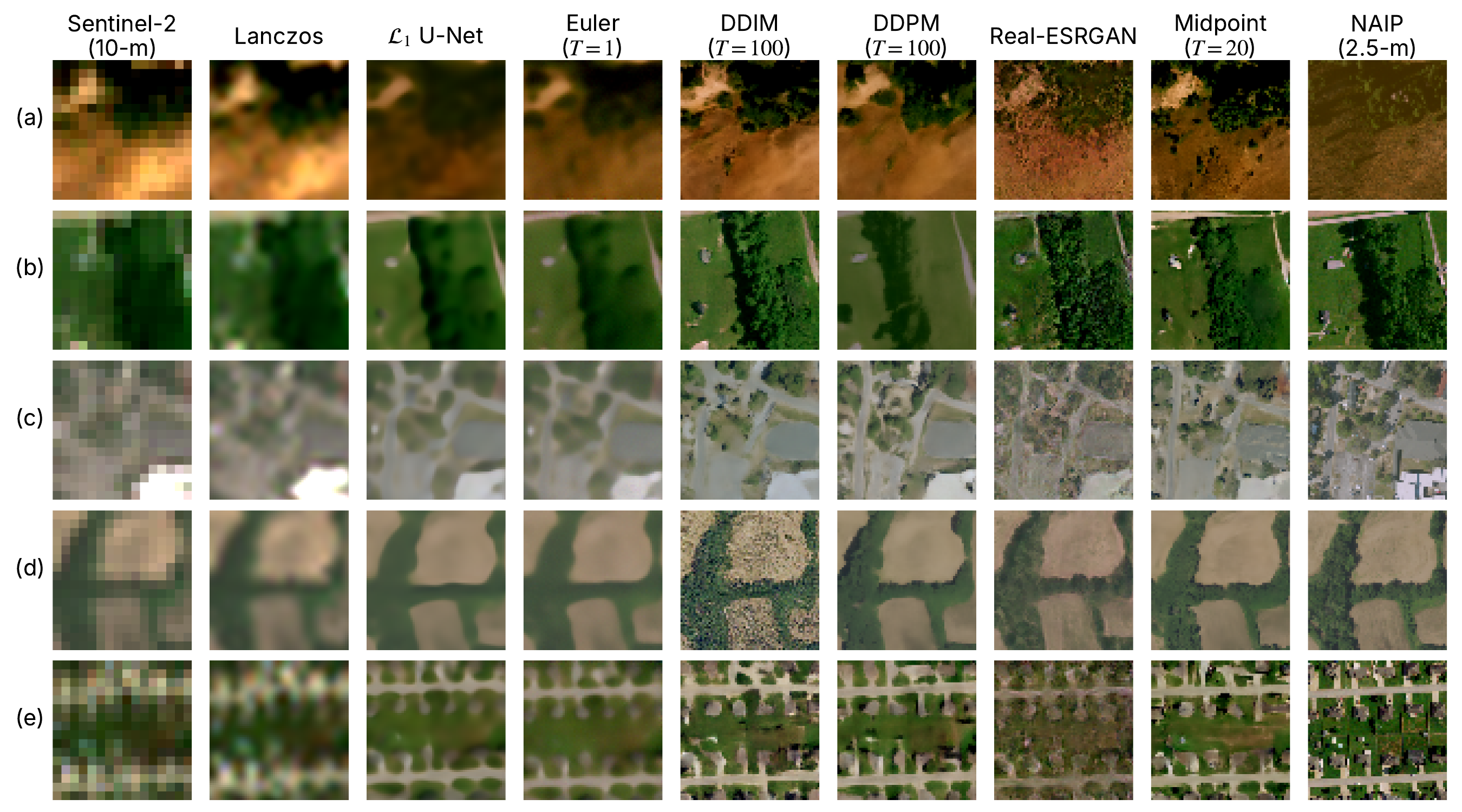}

    \begingroup
    \captionsetup{
    }

    \caption{Visual comparison of super-resolution outputs from a selection of the models evaluated in this study.}\label{fig:visual_comparison}

    \endgroup
\end{figure}

According to the LPIPS metric, Real-ESRGAN produced the most perceptually accurate synthetic imagery with a mean LPIPS of 0.1937, outperforming all other methods, including the best-performing flow matching configurations.
However, examining the super-resolved outputs visually in Figure~\ref{fig:visual_comparison} reveals that the nature of Real-ESRGAN's enhancements is different from those of the flow matching and diffusion-based models.
Notably, the Real-ESRGAN outputs contained noticeable artifacts, including noise and unnatural textures in more complex urban regions with many smaller objects (see samples  c and e in Figure~\ref{fig:visual_comparison}).
Conversely, the flow matching and diffusion models produced much clearer outputs that better preserve the structures of buildings and small objects in these regions.
We attribute this behavior to differences in how the models generate outputs. Real-ESRGAN learns a simple deterministic mapping from a low-resolution to high-resolution space, outputting images that tend to aggregate multiple possible high-resolution solutions.
In contrast, the probabilistic nature of the flow matching and diffusion models enables them to output a single plausible high-resolution sample that better preserves structure without necessarily averaging over multiple possibilities.
The Real-ESRGAN model performed well at reconstructing textures (e.g.,  samples a and b in Figure~\ref{fig:visual_comparison}), but the flow matching model also produced realistic textures without the artifacts observed in the Real-ESRGAN outputs.
 Further, in sample e, the Real-ESRGAN output appears to shift the output color of the grass from a lush green to a muddy brown, while the flow matching and diffusion models better preserve the original spectral characteristics from the low-resolution input.
This spectral shift between the low-resolution input and the Real-ESRGAN output is also present in sample d, where the forest and barren land cover interfaces also appear to have the effect of blurring the forest a shade of brown.
Given that the visual analysis contradicts the findings from the LPIPS metrics, we posit that the LPIPS metric is poorly aligned with human perception of high-resolution remote sensing imagery.
Additionally, as the Real-ESRGAN training includes a perceptual loss based on an ImageNet pre-trained CNN, it should not be surprising that it yields high similarity scores according to LPIPS, which is also based on features extracted from a pre-trained CNN.
As these networks are trained on natural images, they may be blind to remote sensing-specific spectral/spatial features, weakening the utility of the perceptual loss and LPIPS metric for this domain.

\begin{figure}[H]
    \centering
    \includegraphics[width=\textwidth]{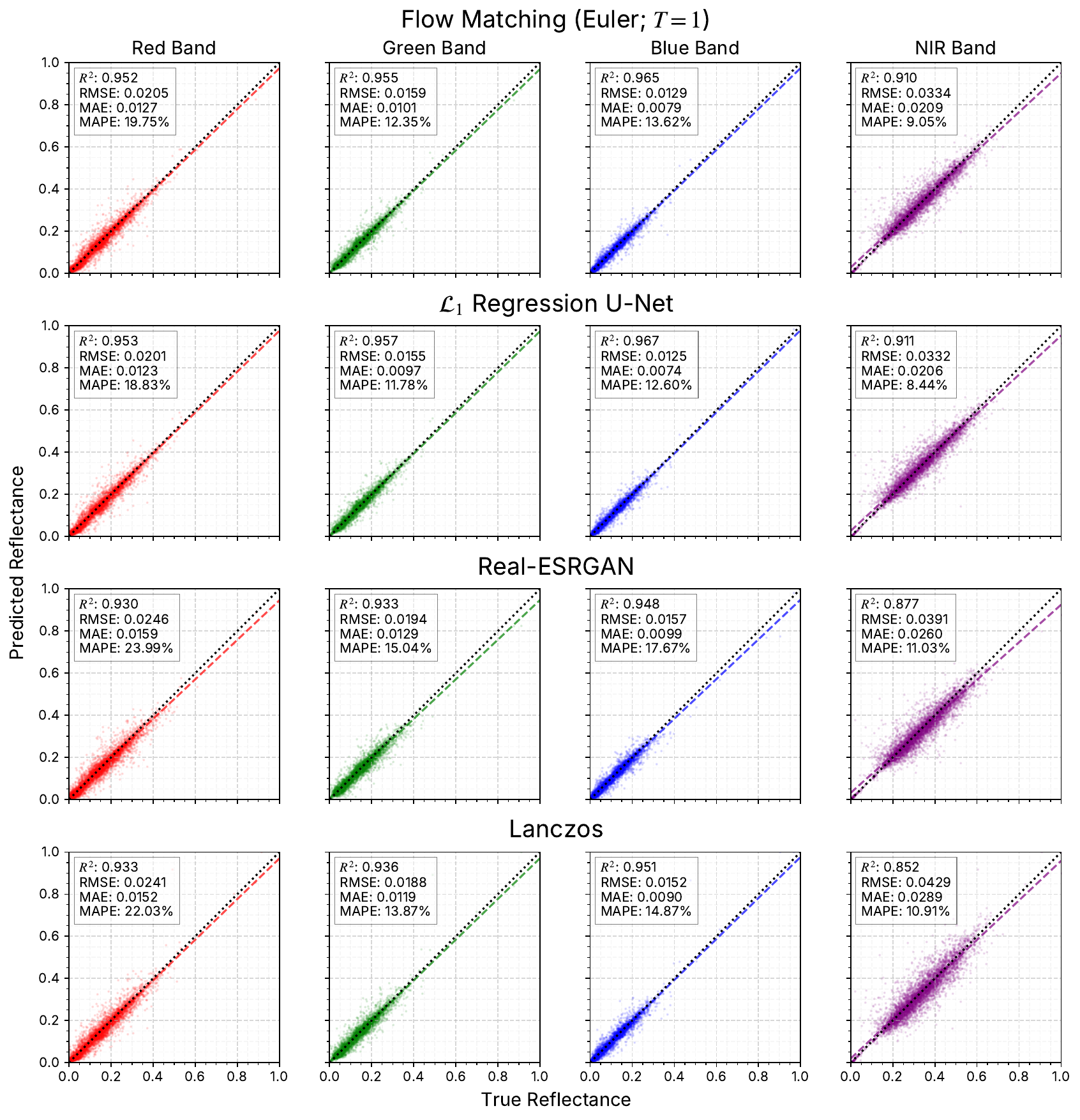}
    \caption{
        Comparison of per-band regression plots for flow matching using the Euler solver with \(T=1\) steps (top), U-Net trained using simple \(\mathcal{L}_1\) Regression (middle-top), Real-ESRGAN (middle-bottom), and Lanczos upsampling (bottom) compared with true high-resolution calibrated NAIP imagery.
    }\label{fig:regression_plots}
\end{figure}

To evaluate the spectral reliability of the super-resolved outputs, we performed a regression analysis where we compared the predicted high-resolution pixel values with the true high-resolution calibrated NAIP pixel values across all pixels in the validation set.
Figure~\ref{fig:regression_plots} shows the per-band regression plots from this analysis, where we compare the outputs of the \(\mathcal{L}_1\) regression U-Net baseline, the flow matching model using the Euler solver at \(T=1\), the Real-ESRGAN model, and Lanczos upsampling.
Both the \(\mathcal{L}_1\) regression U-Net and the flow matching model produced the best fit in terms of predicting the true high-resolution pixel values across all bands, achieving overall \(R^2\) values of 0.969 and 0.968, respectively, compared to 0.955 for Real-ESRGAN and 0.951 for Lanczos upsampling.
Both models similarly achieved the lowest overall error metrics, with overall RMSE values of 0.0218 and 0.0221 and MAE values of 0.0125 and 0.0129 for the \(\mathcal{L}_1\) regression baseline and Euler solver, respectively, compared to RMSE values of 0.0262 and 0.0274 for Real-ESRGAN and Lanczos upsampling.
These results further corroborate the PSNR findings in Table~\ref{tab:sr_results}, demonstrating that the single-step flow matching model closely approaches the pixel-wise accuracy of a dedicated \(\mathcal{L}_1\) regression network, while outperforming all other super-resolution methods (despite lower visual fidelity, as shown in Figures~\ref{fig:fm_var_steps} and~\ref{fig:visual_comparison}).

\begin{figure}[H]
    \centering
    \includegraphics[width=\textwidth]{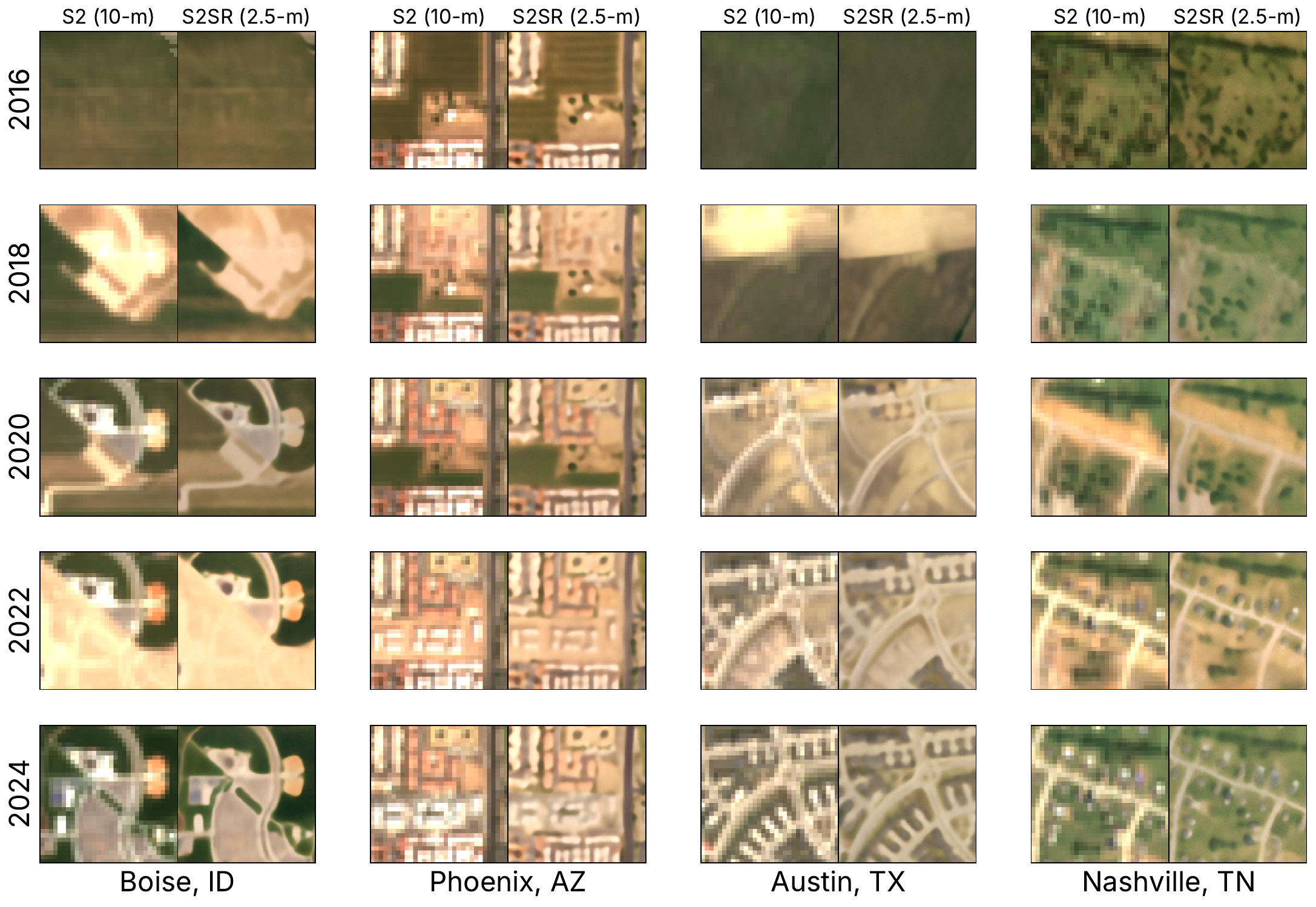}
    \caption{Examples of small-scale urban development visible in the synthetic \SIadj{2.5}{\meter} imagery time series.}\label{fig:temporal_changes}
\end{figure}

From a visual analysis standpoint, even though the Euler solver with \(T=1\) has relatively poor perceptual quality compared to the other solvers with higher values of \(T\), the super-resolved outputs provide substantially more detail in small-scale scenes than their low-resolution Sentinel-2 counterparts.
This can be particularly useful for applications where high-frequency, small-scale features are important, such as urban change analysis.
Figure~\ref{fig:temporal_changes} provides several examples of super-resolved time series showing urban expansion and housing developments in various cities across CONUS, demonstrating the utility of the super-resolved imagery for monitoring fine-scale changes in the landscape over time.
For example, individual structures that are difficult to distinguish in the original Sentinel-2 imagery are easily identifiable in the super-resolved imagery.

\begin{figure}[H]
    \centering
    \includegraphics[width=\textwidth]{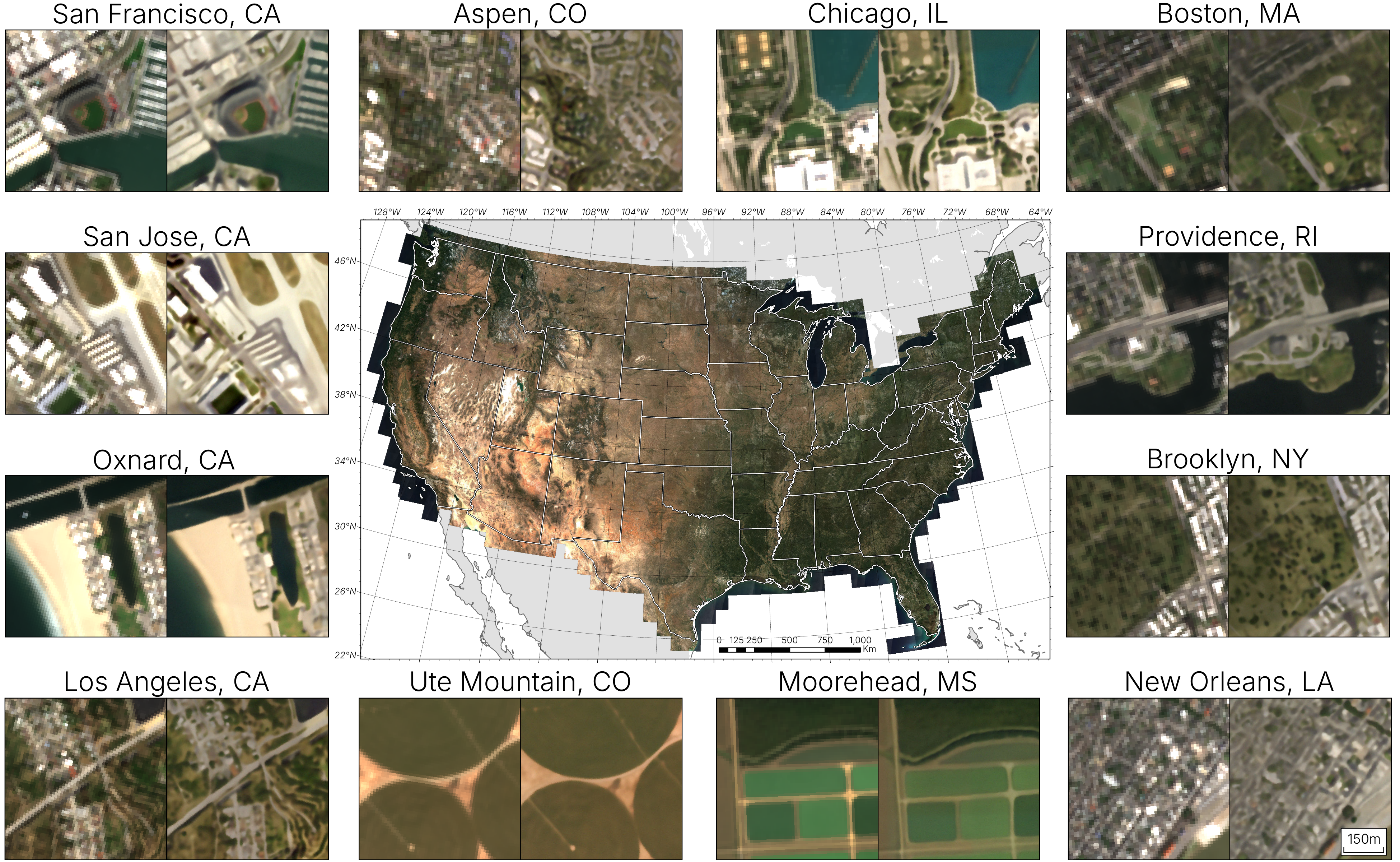}
    \caption{CONUS-wide synthetic \SIadj{2.5}{\meter} imagery for 2025 generated using the flow matching super-resolution method (middle) and selected examples of \SIadj{10}{\meter} Sentinel-2 surface reflectance and synthetic \SIadj{2.5}{\meter} imagery from selected locations across the CONUS (surrounding panels).}\label{fig:conus_sr}

\end{figure}

We used the flow matching model with the Euler solver at \(T=1\) as the core approach for generating the synthetic \SIadj{2.5}{\meter} imagery product for the entire CONUS region for the year 2025 (Section~\ref{sec:methodology:conus_sr}).
A visualization of this product is shown in Figure~\ref{fig:conus_sr}.
In addition to finer delineation of small-scale features in urban environments, the synthetic imagery also improves the delineation of agricultural fields in rural areas (see the examples in Ute Mountain, CO and Moorehead, MS), improving the spatial fidelity of the landscape features across all of CONUS.
We again point out that the model does not hallucinate fine details at \(T=1\), and that the improvement in spatial fidelity is a function of the model refining edges and boundaries between objects and land cover types instead of attempting to recover sub-pixel textures and details.

\subsection{Super-resolution impacts on land cover classification}

\begin{figure}[H]
    \centering
    \includegraphics[width=\textwidth]{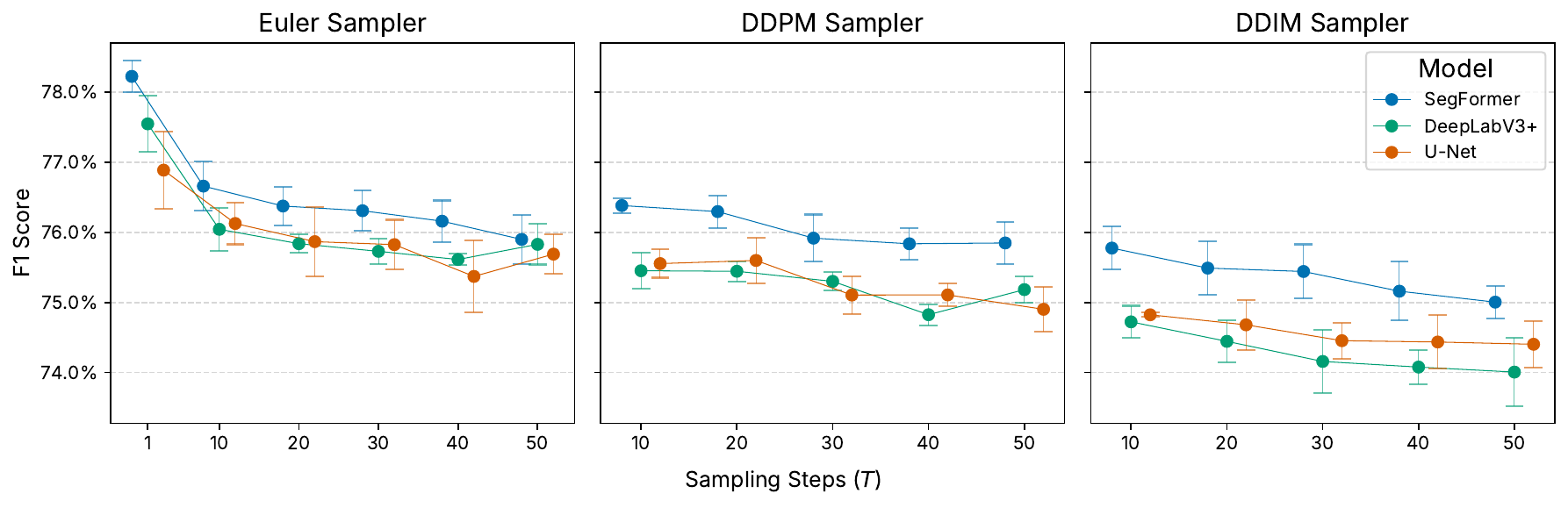}
    \caption{Impact of increasing the number of sampling steps using iterative super-resolution methods on downstream land cover classification F1 Score. The DDIM/DDPM samplers with \(T=1\) have been omitted for clarity due to poor performance. The error bars correspond to the standard deviation.}\label{fig:lc_steps_f1}
\end{figure}

Figure~\ref{fig:lc_steps_f1} shows the performance of the downstream land cover classification models as a function of the number of sampling steps \(T\) across the Euler, DDPM, and DDIM samplers.
Overall, as \(T\) increased, the land cover classification performance tended to decrease across all samplers and segmentation models, similar to the trend in PSNR observed in Figures~\ref{fig:sampling_performance} and~\ref{fig:euler_vs_midpoint}.
This indicates that the increased perceptual quality of the synthetic images at higher values of \(T\) hindered the models' ability to classify land cover types accurately, likely due to the hallucination of fine details that do not necessarily correspond to the true high-resolution imagery, a problematic behavior for remote sensing applications that demand reliable imagery.
 However, using the Euler solver and setting \(T=1\) maintained high classification performance without degrading the reliability of the imagery, and we selected this model configuration for further comparison with Lanczos upsampling, Real-ESRGAN, and the true high-resolution NAIP imagery.

 Overall, the classifiers trained on the deep super-resolution outputs performed similarly to those trained on simple Lanczos upsampling, as shown in Table~\ref{tab:lc_results}.
When considering overall accuracy, the Lanczos upsampling method outperformed both deep super-resolution methods,  but the metric can be misleading in the presence of class imbalance.
We instead emphasize the class-averaged F1 Score metric that penalizes poor class-wise accuracy to provide a more balanced evaluation of model performance.
Using the super-resolved imagery from the Euler solver for classification resulted in a 0.16\% average increase in F1 Score over Lanczos upsampling when averaged across all segmentation architectures, compared to a 0.01\% increase when using the Real-ESRGAN model.
 Performing paired \(t\)-tests on the cross-validation results showed no significant difference between the Euler solver super-resolved imagery and Lanczos upsampling in terms of class-averaged F1 Score, either overall (\(p=0.268\)) or within any individual segmentation architecture after applying the Benjamini--Hochberg correction for multiple comparisons (\(p_{\text{adj}} \ge 0.100\))~\citep{benjaminiControllingFalseDiscovery1995}.
The SegFormer model achieved the highest F1 Score across all super-resolution models, with an average of 78.23\% when using the Euler solver super-resolved imagery, compared to 77.55\% for the DeepLabV3+ model and 76.89\% for the U-Net model when evaluated on the same imagery.
As expected, the high-resolution NAIP imagery provided the best classification performance, achieving an average F1 Score of 80.74\% across all segmentation architectures compared to Euler's 77.56\%, Real-ESRGAN's 77.41\%, and Lanczos' 77.40\%.

\begin{table}[H]
  \centering
  \caption{Comparison of land cover classification metrics across different semantic segmentation architectures and input data sources.}
  \label{tab:lc_results}
  \sisetup{
    separate-uncertainty = true,
    table-format = 2.2(2),
    detect-weight = true,
    mode = text
  }
  \begin{tabular*}{\textwidth}{@{\extracolsep{\fill}}l l S S S S}
    \toprule
    Model & Method/Data & {UA (\%)} & {PA (\%)} & {F1  (\%)} & {OA (\%)} \\
    \midrule
    \multirow{4}{*}{U-Net} & Lanczos & 81.33 \pm 0.68 & 74.19 \pm 0.41 & 76.81 \pm 0.50 & 92.53 \pm 0.23 \\
        & Real-ESRGAN & 81.36 \pm 0.66 & 74.15 \pm 0.20 & 76.75 \pm 0.21 & 92.43 \pm 0.30 \\
        & Euler ($T=1$) & 81.16 \pm 1.23 & 74.49 \pm 0.36 & 76.89 \pm 0.56 & 92.46 \pm 0.18 \\
        & NAIP & 83.85 \pm 0.64 & 78.88 \pm 0.29 & 80.92 \pm 0.21 & 93.81 \pm 0.05 \\
    \addlinespace
    \multirow{4}{*}{DeepLabV3+} & Lanczos & 81.77 \pm 0.37 & 74.80 \pm 0.31 & 77.45 \pm 0.21 & 92.79 \pm 0.01 \\
        & Real-ESRGAN & 82.14 \pm 0.30 & 74.58 \pm 0.35 & 77.35 \pm 0.33 & 92.69 \pm 0.02 \\
        & Euler ($T=1$) & 82.40 \pm 0.55 & 74.79 \pm 0.41 & 77.55 \pm 0.40 & 92.71 \pm 0.02 \\
        & NAIP & 83.65 \pm 0.13 & 78.09 \pm 0.18 & 80.38 \pm 0.15 & 93.70 \pm 0.02 \\
    \addlinespace
    \multirow{4}{*}{SegFormer} & Lanczos & 81.15 \pm 0.38 & 75.82 \pm 0.30 & 77.94 \pm 0.27 & 92.93 \pm 0.02 \\
        & Real-ESRGAN & 81.74 \pm 0.44 & 75.80 \pm 0.21 & 78.12 \pm 0.24 & 92.82 \pm 0.01 \\
        & Euler ($T=1$) & 81.90 \pm 0.14 & 75.90 \pm 0.26 & 78.23 \pm 0.23 & 92.88 \pm 0.01 \\
        & NAIP & 83.43 \pm 0.31 & 79.09 \pm 0.16 & 80.92 \pm 0.05 & 93.84 \pm 0.01 \\
    \bottomrule
  \end{tabular*}
\end{table}

\begin{figure}[H]
    \centering
    \includegraphics[width=\textwidth]{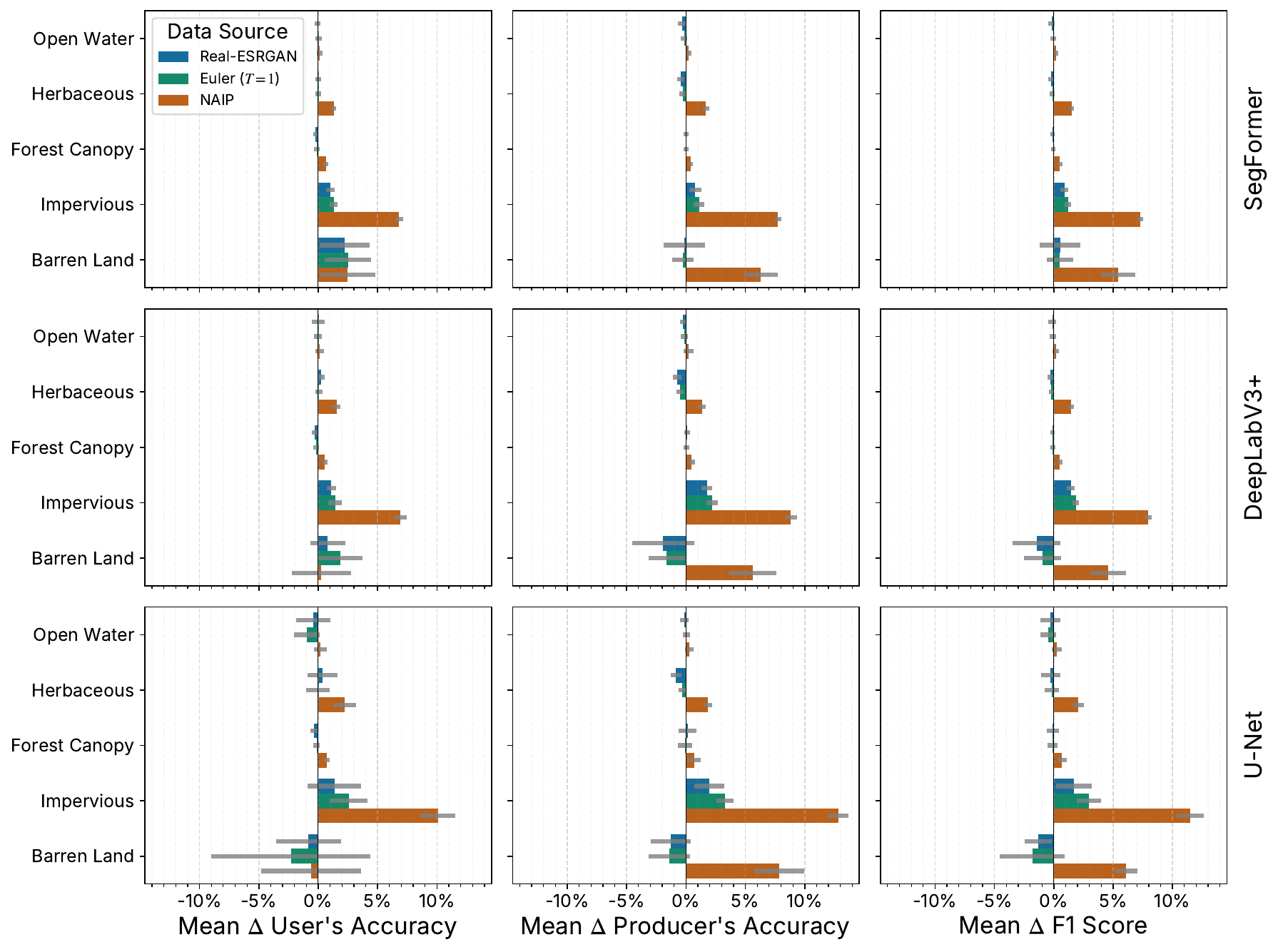}
    \caption{Per-class changes in accuracy metrics compared to Lanczos-upsampled Sentinel-2 imagery for super-resolution land cover classification when using imagery from Real-ESRGAN, our flow matching (Euler solver with \(T=1\)) model, and NAIP imagery. Error bars correspond to the standard deviation.}\label{fig:lc_class_grid}
\end{figure}

While the overall differences between the deep super-resolution methods were small compared to Lanczos upsampling, examining the class-wise change in metrics provides further insight into the benefits and drawbacks of the deep super-resolution methods.
Figure~\ref{fig:lc_class_grid} shows the per-class changes in classification metrics across each segmentation architecture when moving from Lanczos upsampling to the Real-ESRGAN, flow matching (Euler solver at \(T=1\)), and NAIP imagery.
 When averaged across all segmentation architectures, the deep super-resolution methods tended to improve classification of the impervious class, with the flow matching method yielding a statistically significant 2.00\% increase in F1 Score (\(p_{\text{adj}} < 0.001\)) and Real-ESRGAN providing a 1.34\% increase.
This is offset by a decrease in performance for the barren land class, where using the super-resolved data resulted in 0.75\% decreases in F1 Score for both deep super-resolution methods compared to Lanczos upsampling , though this change was not statistically significant for the flow matching model (\(p_{\text{adj}} = 0.133\)).

\begin{figure}[H]
    \centering
    \includegraphics[width=\textwidth]{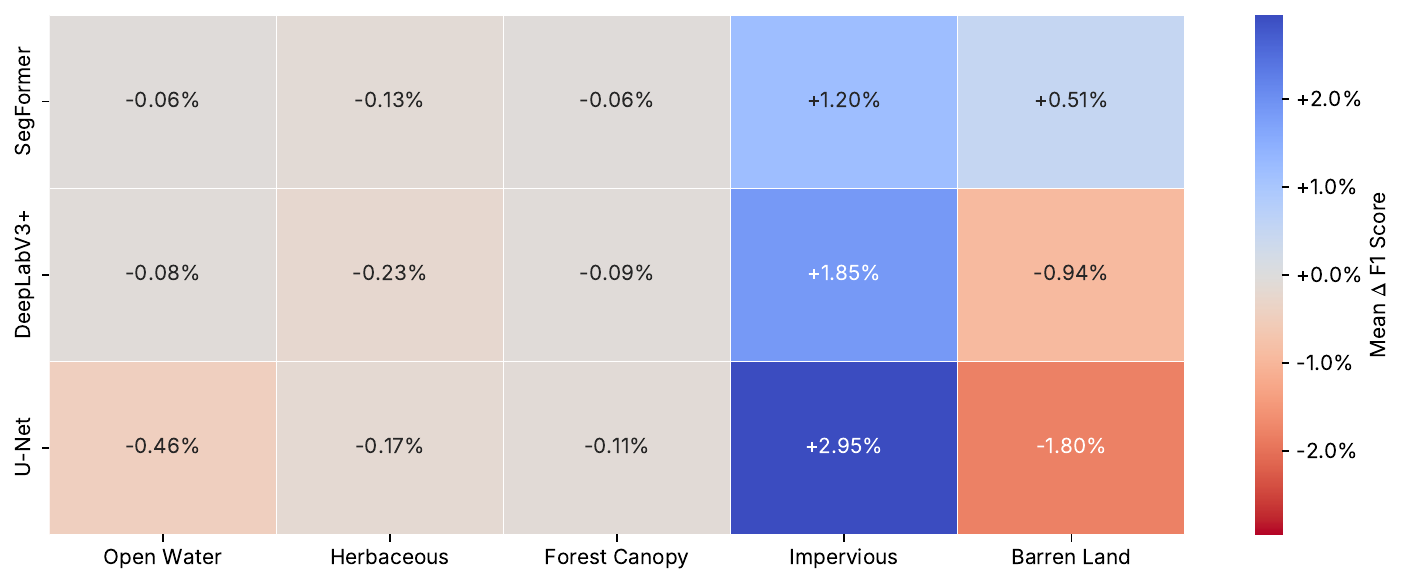}
    \caption{Mean per-class changes in F1 Score for our flow matching model (Euler solver with \(T=1\)) compared to Lanczos-upsampled Sentinel-2 imagery for super-resolution land cover classification when using SegFormer (top row), DeepLabV3+ (middle row), and U-Net (bottom row)}\label{fig:lc_class_heatmap}
\end{figure}

\begin{figure}[H]
    \centering
    \includegraphics[width=\textwidth]{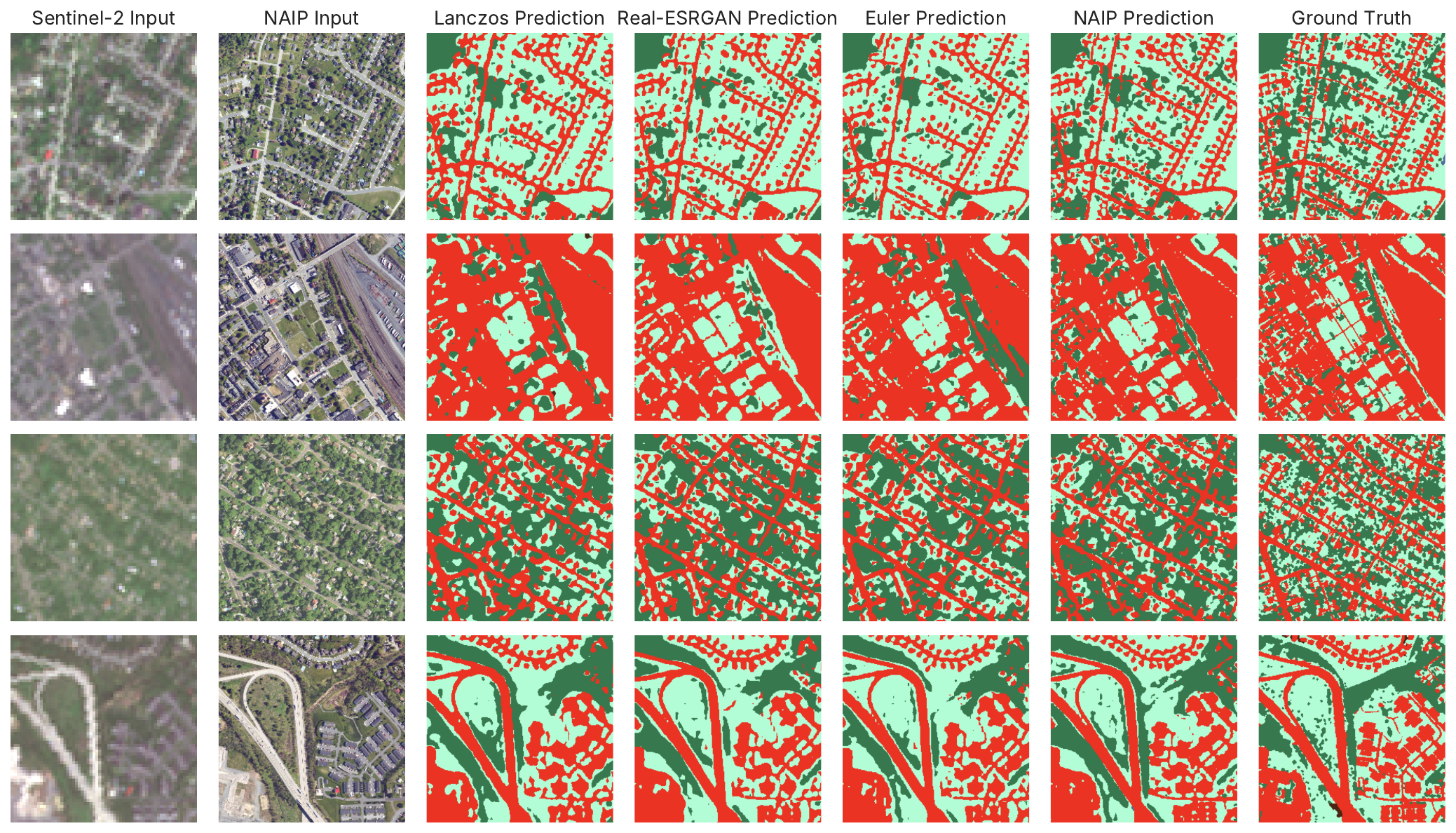}
    \caption{Comparison of urban land cover classification outputs from SegFormer at 2.5-m using different super-resolution methods for upsampling Sentinel-2 imagery.}\label{fig:lc_model_comparison}
\end{figure}

 Figure~\ref{fig:lc_class_heatmap} shows that the magnitude of these per-class shifts in F1 Score is also architecture-dependent.
The U-Net model benefited the most from the flow matching imagery when classifying impervious surfaces (+2.95\%), while the SegFormer model saw the smallest improvement (+1.20\%).
Concurrently, there was a reverse trend with the barren land class, where the U-Net model performed worse with the super-resolved imagery (-1.80\%) while the SegFormer model showed a slight increase (+0.51\%).
Across all architectures, the changes in performance for the natural land cover classes were not significant (\(p_{\text{adj}} > 0.05\)).
Both the impervious and barren classes showed the largest discrepancy in F1 Score between the models trained using some form of Sentinel-2 data and the models trained using true high-resolution NAIP imagery, with the NAIP-trained models achieving 7.80\% and 5.86\% higher F1 Scores for the impervious and barren classes, respectively, compared to the average of all Sentinel-2 based methods (including Lanczos upsampling).
 These improvements in classifying impervious surfaces are visualized in Figure~\ref{fig:lc_model_comparison}, where the models trained using the super-resolved imagery tend to delineate roads and buildings slightly better compared to the model trained on Lanczos-upsampled imagery.

\begin{figure}[H]
    \centering
    \includegraphics[width=\textwidth]{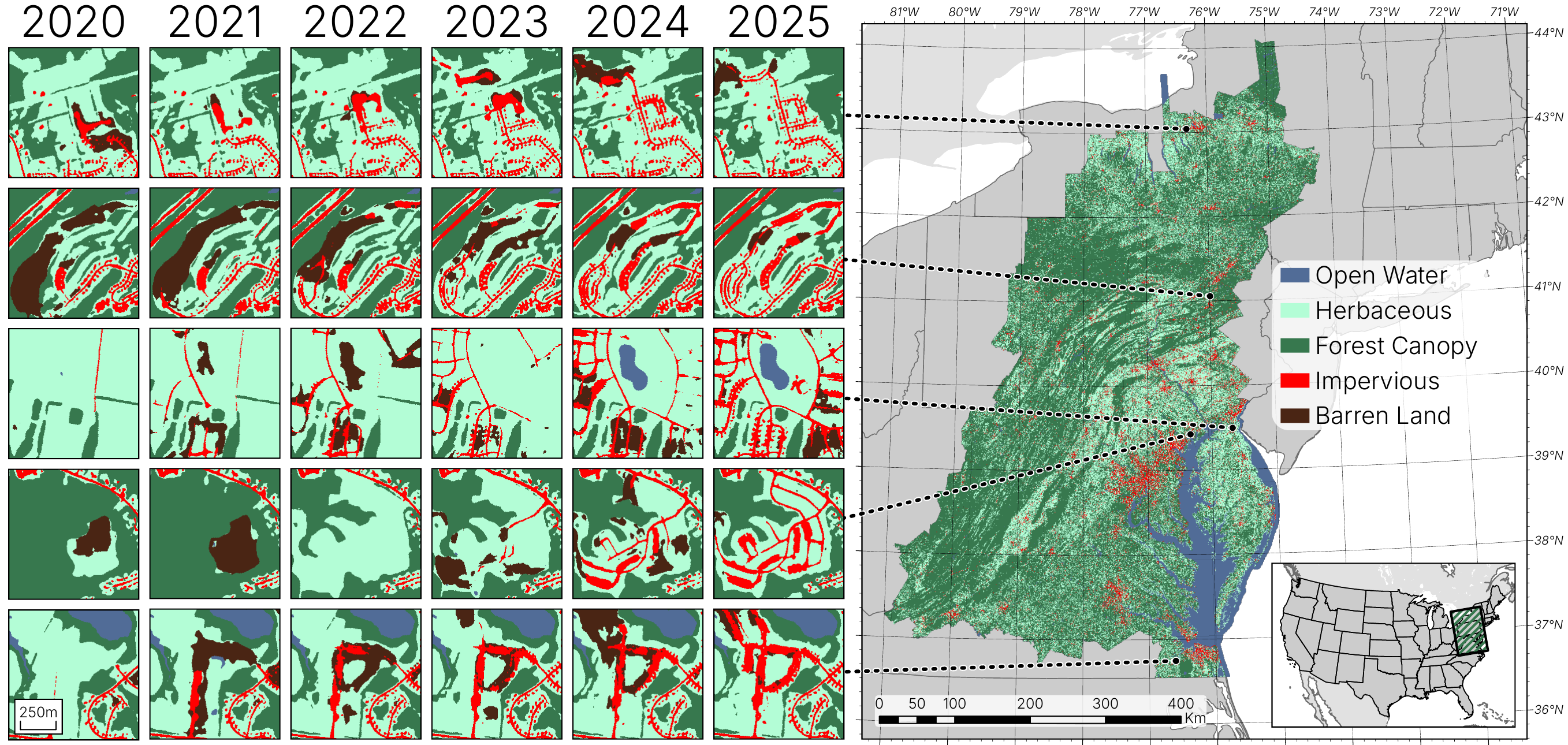}
    \caption{Examples of high-frequency land cover changes captured by our annual Chesapeake Bay \SIadj{2.5}{\meter} land cover product.}\label{fig:lc_temporal}
\end{figure}

We applied the SegFormer model trained using the flow matching super-resolved imagery from the best-performing run of the cross-validation procedure (in terms of F1 Score) to the Chesapeake Bay region to generate a \SIadj{2.5}{\meter} land cover map for the region following the procedure described in Section~\ref{sec:methodology:lc}.
The results of the accuracy assessment of  the 2021 and 2022 years of our land cover map against the ground truth assessment points from the CBLC dataset are presented in Table~\ref{tab:lc_assessment}.
Overall, our product achieved an overall accuracy of 89.11\% and a class-averaged F1 Score of 74.40\% over the 25,000 assessment points, with particularly strong performance for the open water and forest canopy classes with F1 Scores of 97.18\% and 92.51\%, respectively.
The impervious class was classified fairly well, with an F1 Score of 70.52\%, with a slightly higher producer's accuracy (72.38\%) than user's accuracy (68.74\%), indicating that the impervious class is slightly more prone to errors of commission than omission.
Barren lands were poorly classified, with an F1 Score of 27.85\%, and a producer's accuracy of 18.64\% compared to a user's accuracy of 55.00\%, indicating a high tendency for errors of omission.
However, this had little impact on the overall accuracy, as only 118 of the 25,000 assessment points were labeled as barren land in the ground truth dataset (0.47\%).
The confusion matrix reveals misclassifications between the herbaceous and barren classes, with 59 of the 118 barren land points being misclassified as herbaceous.
Therefore, despite a high overall accuracy and moderate to high F1 Scores for the other classes, the performance of the model for classifying barren land is poor and should be interpreted with caution when using this product for applications where barren land is of interest.
We note that the original CBLC dataset estimates that the producer's and user's accuracy for the barren land class is 40\% and 63\%, respectively; we therefore partially attribute the poor performance of our model for classifying barren land to the general difficulty of classifying this class in the ground truth dataset~\citep{mcdonaldChesapeakeBayLand2025}.
A full visualization of the land cover product for 2025 is shown in Figure~\ref{fig:lc_temporal}, along with several examples of rapid land cover change captured by the product to demonstrate the utility of this dataset for monitoring fine-scale changes in the landscape over time.
 We note that these results were only validated for the 2021 and 2022 years for which ground truth data was available from the CBLC dataset.
These results should be interpreted cautiously when extrapolating across the full temporal scope of the product.

\begin{table}[H]
    \centering
    \caption{Confusion matrix for annual \SIadj{2.5}{\meter} Chesapeake Bay land cover product (2021 and 2022).}\label{tab:lc_assessment}
    \footnotesize
    \begin{tabular*}{\textwidth}{@{\extracolsep{\fill}}lrrrrrrr}
        \toprule
         & \multicolumn{5}{c}{\textbf{Predicted Class}} & & \\
        \cmidrule(lr){2-6}
        \textbf{True Class} & Open Water & Herbaceous & Forest Canopy & Impervious & Barren Land & \textbf{Total} & \textbf{PA (\%)} \\
        \midrule
        Open Water & \textbf{3,324} & 41 & 23 & 5 & 1 & 3,394 & 97.94 \\
        Herbaceous & 84 & \textbf{6,245} & 644 & 288 & 10 & 7,271 & 85.89 \\
        Forest Canopy & 25 & 941 & \textbf{11,648} & 166 & 3 & 12,783 & 91.12 \\
        Impervious & 4 & 319 & 69 & \textbf{1,038} & 4 & 1,434 & 72.38 \\
        Barren Land & 10 & 59 & 14 & 13 & \textbf{22} & 118 & 18.64 \\
        \midrule
        \textbf{Total} & 3,447 & 7,605 & 12,398 & 1,510 & 40 & \textbf{25,000} & 73.20 \\
        \textbf{UA (\%)} & 96.43 & 82.12 & 93.95 & 68.74 & 55.00 & 79.25 & OA: 89.11 \\
        \bottomrule
    \end{tabular*}
\end{table}

\section{Discussion and conclusion}\label{sec:discussion}

Modeling techniques for deep super-resolution of remote sensing imagery are rapidly evolving, and diffusion-based methods have begun to address several limitations of traditional GAN-based approaches~\citep{liuDiffusionModelsMeet2024}.
However, diffusion models often require large numbers of sampling steps to produce high-quality outputs that are comparable to their GAN-based counterparts, leading to slow inference times that limit their use in large-scale applications~\citep{woltersZoomingOutZooming2023}.
In this work, we showed that flow matching-based super-resolution models are both more performant and more computationally efficient than diffusion-based approaches for the task of 4\(\times\) Sentinel-2 super-resolution.
Our flow matching model outperformed both the Real-ESRGAN and diffusion models in terms of both pixel-wise similarity (Table~\ref{tab:sr_results}, Figure~\ref{fig:regression_plots}) and qualitative visual fidelity (Figure~\ref{fig:visual_comparison}).
Beyond image similarity, we demonstrated that our flow matching model is reliable for downstream land cover classification tasks, outperforming simple Lanczos upsampling, Real-ESRGAN, and diffusion-based super-resolution methods when applied to a land cover mapping task using various semantic segmentation models (Figure~\ref{fig:lc_steps_f1}, Table~\ref{tab:lc_results}).
While the overall improvements in classification performance were relatively small in magnitude, they were more pronounced when examining per-class metrics (Figure~\ref{fig:lc_class_grid}), which showed  statistically significant increases in impervious land cover classification accuracy when using the flow matching super-resolution imagery as input to the land cover classification models compared to Lanczos upsampling.

One key advantage of the flow matching model was its ability to produce outputs that minimize distortion while also being able to provide high perceptual quality depending on the choice of solver used and the number of sampling steps employed during inference.
Figures~\ref{fig:euler_vs_midpoint} and~\ref{fig:fm_var_steps} illustrate how the flow matching model produced images with high accuracy in only a single step with the Euler solver, while also achieving high perceptual quality when using the Midpoint solver with more sampling steps.
While GAN-based models can be trained to emphasize perceptual quality over distortion (or vice versa) by adjusting the relative importance of the pixel-wise loss, adversarial loss, and perceptual loss during training, the model's behavior is fixed once trained, and a user has no control over this trade-off at inference time~\citep{blauPerceptionDistortionTradeoff2018}.
For remote sensing applications, this property is particularly useful, as imagery with high perceptual quality may be preferred for human interpretation tasks, while imagery with high pixel-wise accuracy is more suitable for analytical workflows~\citep{aybarRadiometricallySpatiallyConsistent2026}.

For the Real-ESRGAN baseline, we adopted the perceptual and adversarial loss weights recommended in the original study~\citep[][\(\lambda_{\text{percep}}=1.0$ and $\lambda_{\text{adv}}=0.1\)]{wangRealESRGANTrainingRealWorld2021}.
Because these weights determine where Real-ESRGAN sits on the perception-distortion trade-off, other settings would shift its balance between fidelity and perceptual quality, though each adjustment requires retraining.
In contrast, the flow matching model can navigate this trade-off at inference time by varying the number of integration steps.
While a key advantage of our flow matching approach is that it does not rely on ancillary networks, such losses are also not exclusive to GAN-based methods.
By extrapolating a clean sample at each training step via $\hat{x}_{1}=x_{t}+\left(1-t\right)v_{\theta}\left(x_{t},t\right)$ for flow matching, or $\hat{x}_{0}=\frac{1}{\sqrt{\bar{\alpha}_{t}}}\left(x_{t}-\sqrt{1-\bar{\alpha}_{t}}\,\epsilon_{\theta}\left(x_{t},t\right)\right)$ for diffusion, the prediction can be passed directly to a discriminator or feature extraction network.
Subsequently, similar hyperparameter tuning can be applied to the flow matching and diffusion models to adjust their behavior during training.
The comparisons between Real-ESRGAN and the diffusion-based or flow matching-based models should therefore be interpreted in the context of the specific training recipe adopted in our study.

A similar consideration around our experimental design involves the use of \(\mathcal{L}_1\) loss as opposed to \(\mathcal{L}_2\) loss.
As discussed in Sections~\ref{sec:methodology:sr:unet_baseline} and~\ref{sec:methodology:sr:training}, we opted for \(\mathcal{L}_1\) loss to train our models as it has been shown to provide clear benefits for super-resolution tasks over standard \(\mathcal{L}_2\) loss~\citep{zhaoLossFunctionsImage2017, anagunSRLibraryComparingDifferent2019, limEnhancedDeepResidual2017}.
Even for diffusion super-resolution models, \(\mathcal{L}_1\) provides tangible benefits in terms of output perceptual similarity compared to \(\mathcal{L}_2\), as demonstrated by~\citet{sahariaImageSuperResolutionIterative2021} and adopted by~\citet{liSRDiffSingleImage2022} and~\citet{xiaDiffIREfficientDiffusion2023} in their SRDiff and DiffIR techniques for super-resolving with diffusion, respectively.
However, this introduces incongruities with the original formulations of the diffusion and flow matching theoretical foundations.
In the DDPM formulation, \(\mathcal{L}_2\) loss is used to minimize the variational bound on negative log-likelihood, which decomposes into Kullback-Leibler divergences between forward and reverse Gaussian transitions.
These transition distributions share isotropic Gaussian covariances, and as such this bound reduces to a squared Euclidean (\(\mathcal{L}_2\)) distance~\citep{hoDenoisingDiffusionProbabilistic2020}.
While this departs from the formal theoretical derivations of DDPM, adopting an \(\mathcal{L}_1\) objective represents a well-established practical compromise in super-resolution modeling, where empirical visual fidelity is prioritized over strict faithfulness to the formal mathematical foundations.
Similarly, under flow matching, the proof that Conditional Flow Matching objectives converge to the true marginal vector field depends on \(\mathcal{L}_2\) regression recovering the conditional mean of the velocity~\citep{lipmanFlowMatchingGenerative2023}, while \(\mathcal{L}_1\) loss encourages the model to learn the conditional median instead.
For the flow matching model, we used \(\mathcal{L}_1\) such that our proposed approach is more comparable with the other super-resolution techniques, which all utilize the same objective.

This choice connects our flow-matching model directly to the supervised regression baseline.
When applying the Euler solver with \(T=1\) to the flow matching model at \(t=0\), evaluating the Euler step under an \(\mathcal{L}_1\) objective effectively translates to \(\operatorname{median}(x_1 \mid x_{\text{LR}})\), approximating the outputs of our supervised \(\mathcal{L}_1\) U-Net baseline in terms of pixel-wise metrics, as shown in Table~\ref{tab:sr_results} and Figure~\ref{fig:regression_plots}.
We attribute this small difference (overall \(R^2\) of 0.968 for Euler vs. 0.969 for the baseline) to parameter allocation: the regression U-Net uses its entire capacity to learn a static mapping from a low-resolution input to a high-resolution output, whereas the flow matching network must distribute its capacity across the continuous interval \(t \in [0, 1)\) to model the complete velocity field.
In return for this minor trade-off at \(T=1\), the flow matching framework provides the flexibility to navigate into perceptually detailed generative sampling simply by evaluating additional integration steps.
Even at \(T=1\), the flow matching model achieved a lower LPIPS than the regression baseline (0.3738 vs. 0.4437), demonstrating moderately improved perceptual quality despite comparable spectral distortion.

To evaluate the utility of the super-resolved imagery, we trained land cover classifiers using super-resolved imagery from various methods as input.
Imagery from the single-step Euler solver performed comparably to Lanczos upsampling overall and significantly improved F1 Score for impervious land cover classification.
One interpretation of this result is that the super-resolved imagery acts as a learned prior on sub-pixel structure, which segmentation models can use to improve predictions of fine-grained surface features (e.g., roads, buildings, and other small-scale structures).
If so, the benefit should be larger when training data are limited, because the classifier no longer has to learn these features from upsampled low-resolution imagery alone.
Future investigations should evaluate the impact of super-resolved imagery in downstream tasks, particularly where such tasks revolve around resolving small structures or phenomena that are finer than the native resolution of the low-resolution input imagery or involve analysis of complex scenes with limited data.
Such explicit pre-processing may be particularly advantageous for workflows that cannot rely on deep spatial architectures to recover fine geometry, such as manual photointerpretation or traditional statistical and pixel-wise classifiers that operate without spatial context.

The use of composite imagery to create the synthetic imagery product as opposed to standard Sentinel-2 imagery does impose potential limitations on the observed performance of the model due to domain drifts between the two imagery types.
While temporal composites are helpful from an operational and data consistency standpoint, they pose challenges when applied to models trained on raw single-date imagery.
Annual mean composites have different spatial and spectral characteristics compared to single acquisitions caused by suppressed sensor noise, an altered effective point spread function due to multi-angle aggregation, and temporally averaged surface reflectance across seasonal changes.
Given that there is no publicly available source from which to draw \SIadj{2.5}{\meter} temporal composites with expected spectral characteristics similar to annual surface reflectance composites, we cannot provide a robust quantitative evaluation of the impact of this domain shift.
However, when examining the outputs of the model visually (Figures~\ref{fig:temporal_changes} and~\ref{fig:conus_sr}), we see that the resulting super-resolutions are consistent with the outputs of the flow matching model evaluated with the Euler solver and \(T=1\) on single-acquisition imagery in Figures~\ref{fig:fm_var_steps} and~\ref{fig:visual_comparison}.
Given that the Euler solver with \(T=1\) has reduced generative capabilities and does not have a tendency to hallucinate high-frequency features, this likely serves as a kind of regularization, sharpening edges without inventing complex sub-pixel structures that would likely be absent from the temporal composite.

 While pixel-wise metrics cannot be computed between the super-resolved outputs and a concurrent \SIadj{2.5}{\meter} aerial composite ground truth (as NAIP imagery represents instantaneous, single-date snapshots rather than multi-temporal composites), our downstream regional land cover assessment serves as an end-to-end task-specific validation of this domain shift.
The fact that SegFormer achieved an overall accuracy of 89.11\% across 25,000 independently sampled ground-truth points over the Chesapeake Bay watershed (Table~\ref{tab:lc_assessment}) demonstrates that the super-resolved annual composites still result in robust spatial and radiometric fidelity for downstream analysis.
Still, the Chesapeake Bay land cover data product may also be affected by this drift in input imagery type. While the classifier achieves relatively high F1 scores across most classes in the final product, the weak barren performance in the temporal composite likely arises as a result of the classifier being affected by this distribution shift.
The poor producer’s accuracy (18.64\%) indicates that the temporal composites likely have more vegetation, which can vary across the season: since the NAIP imagery used to generate the ground truth labels is acquired at a single point in time, the ground at a given location may appear barren later in the year (or after harvest in the case of agricultural fields), but has bright vegetation at other points in the year, skewing the mean reflectance higher.
This could be an explanation for the high level of confusion between the herbaceous and barren land cover classes in the final land cover product.
We do note that our land cover classification training dataset suffered from an obvious spatial bias away from the dense urban areas of the Delaware and Baltimore-Washington D.C. corridors present in the Chesapeake Bay region.
Figure~\ref{fig:cpb_samples} shows that the sampled locations are concentrated in more rural and natural areas, instead of being uniformly distributed across the entire region.
We attribute this to our strict quality control measures and narrow temporal window for acquiring cloud-free Sentinel-2 and NAIP imagery, restricting the number of regions that met the criteria for inclusion in our dataset.
The point-based assessment dataset was not constrained by the need for temporal alignment between the Sentinel-2 and NAIP imagery, enabling the assessment of a larger number of points across the region where training data was absent when evaluating the final land cover product.
 However, the point-based assessment consisted only of data from 2021 and 2022, and we do not quantify the accuracy of the other temporal periods.
We anticipate that the accuracy might extrapolate to the rest of the dataset as the Sentinel-2 sensor characteristics did not meaningfully change for the 2020 or 2023--2025 periods compared to 2021 and 2022.

One potential limitation of our synthetic imagery dataset is a noticeable shift in brightness in the outputs compared to the Sentinel-2 surface reflectance inputs, as shown in Figures 11 and 12.
This is likely due to the difference in the characteristics of the Sentinel-2 imagery and the NAIP imagery.
The Sentinel-2 data we used as input is an atmospherically-corrected surface reflectance product, whereas the NAIP imagery is a simple imagery product that uses digital numbers that do not correspond to meaningful physical quantities.
Our cross-calibration attempts to correct for this in the training data by building a linear model between the aligned Sentinel-2 and NAIP pixels of each chip.
However, as the NAIP imagery is downsampled to enable this, this may introduce some distortion into the spectral makeup of the chips.
For instance, given that the point spread function is different between the Sentinel-2 and NAIP sensors, small, bright objects (such as buildings) appear to have a much smaller footprint in the NAIP imagery compared to the Sentinel-2 imagery.
After downsampling, those bright regions become under-represented in the pixel values used for calibration, biasing the model.
Further, given that the model is a simple ordinary least squares regression, it is unable to capture non-linear relationships between the NAIP imagery’s digital numbers and the Sentinel-2 surface reflectance.
This apparent shift in brightness can be observed in the ground truth samples in Figure~\ref{fig:s2naip_samples}, as well as the outputs of the other super-resolution models we tested (as shown in Figure~\ref{fig:visual_comparison}, sample c).
Further, we do not observe any substantial biases in the RGB bands in Figure~\ref{fig:regression_plots}, indicating that the models we tested are faithfully matching the target pixel brightness distributions of the cross-calibrated NAIP images.
Future work should consider more advanced calibration routines: we hypothesize that non-linear models (such as histogram matching) applied across full Sentinel-2 scenes and full-resolution NAIP orthoimage tiles could improve this cross-calibration procedure, reducing this brightness shift compared to our simpler linear chip-based approach.

As part of our analysis of flow matching super-resolution, we investigated the impact of different fixed-step ODE solvers on sampling performance.
We found that at low numbers of sampling steps (e.g., \(T \leq 10\)), the first-order Euler solver and second-order Midpoint solver outperformed the second-order Heun and fourth-order RK4 solvers in terms of both pixel-wise accuracy and perceptual quality (Figure~\ref{fig:euler_vs_midpoint}).
While the Heun and RK4 solvers eventually exceeded the performance of the Euler and Midpoint solvers at higher numbers of sampling steps, the computational cost of these solvers was substantially higher than the simpler Euler and Midpoint methods, making them less attractive for practical use cases (Figure~\ref{fig:sampling_performance}).
We limited our methodology to fixed-step solvers in this work, but adaptive-step solvers~\citep[such as the Dormand-Prince method;][]{dormandFamilyEmbeddedRungeKutta1980} may provide better performance by allocating more steps towards regions of high complexity in the flow field while using fewer steps in regions of low complexity, potentially improving sampling efficiency and output quality.

Beyond the choice of ODE solver for the flow matching models, both the diffusion and flow matching models rely on a myriad of design choices that influence behavior.
For example, the canonical noise prediction parameterization of the diffusion models was used to align with the DDPM and DDIM formulations.
This choice of parameterization could explain the poor performance of the DDPM and DDIM models when evaluated using \(T=1\), as the model is tasked with predicting noise at \(t=T\), where the input consists of pure Gaussian noise and \(\bar{\alpha}_T \approx 0\).
Estimating the clean image requires scaling \(\epsilon_\theta\) by coefficients determined by the variance schedule (Equations~\ref{eq:ddpm_model_train} and~\ref{eq:ddim_reverse}), which are relatively extreme when striding across the entire schedule in a single discrete step, potentially leading to amplified errors.
One could train a flow matching model using the diffusion probability paths, though~\citet{lipmanFlowMatchingGenerative2023} showed that using simple optimal transport displacement trajectories resulted in better performance compared to curved diffusion paths.
Furthermore, different trajectories can be used in flow matching models to increase the efficiency of the sampling procedure or improve the fidelity of the generated outputs, such as 2-Rectified Flow~\citep{liuFLOWSTRAIGHTFAST2023} or mini-batch optimal transport~\citep{pooladianMultisampleFlowMatching2023,tongImprovingGeneralizingFlowbased2024}.

During evaluation, we observed that the LPIPS metric sometimes did not fully align with the perceived visual quality of the super-resolved images.
For example, while the Real-ESRGAN outputs had lower LPIPS values by a wide margin compared to the flow matching results (Table~\ref{tab:sr_results}), their super-resolutions were of lower visual quality (Figure~\ref{fig:visual_comparison}).
As LPIPS uses a convolutional feature extractor pre-trained on general-purpose images from ImageNet, we hypothesize that these features are less relevant for remote sensing imagery, leading to a disconnect between the LPIPS scores and actual visual quality.
Since the Real-ESRGAN model is trained to minimize a perceptual loss based on similar convolutional features, we believe that this led to the Real-ESRGAN outputs being optimized for low LPIPS scores, producing images that were similar to ground truth in terms of the extracted features, but not necessarily visually realistic.
 In fact, using these pre-trained networks for perceptual loss calculation or similarity evaluation on the multispectral imagery necessitated the use of a PCA-based transformation to convert the 4-band model outputs and NAIP imagery into 3-channel RGB images, potentially causing further misalignment between the features extracted from these images and the actual content of the multispectral imagery.
This misalignment could be a contributing factor to the observed split between the LPIPS metric and perceived similarity in our study.
This highlights the need for better perceptual similarity metrics and loss functions that are specific to remote sensing imagery.
 This is one potential use for remote sensing foundation models~\citep{xiaoFoundationModelsRemote2025}, which could serve as deep feature extractors that are well-aligned with the unique characteristics of remote sensing imagery, though early efforts in this direction have shown mixed results~\citep{aybarRadiometricallySpatiallyConsistent2026}.

As flow matching is very similar to diffusion modeling in that both frameworks use a neural network to transform a simple distribution into a more complex data distribution via a series of steps, many of the extensions and improvements that have been proposed for diffusion models may also be applicable to flow matching.
While we utilized SR3-style conditioning with a U-Net architecture in this work for simplicity and easier comparison to existing diffusion-based methods, adapting SRDiff (and its remote sensing derivatives, EDiffSR~\citep{xiaoEDiffSREfficientDiffusion2024} and FastDiffSR~\citep{mengConditionalDiffusionModel2024}) to the flow matching framework is straightforward and may yield further performance improvements.
One advantage of SR3-style conditioning is that adding additional spatial conditioning inputs (e.g., land cover maps, road/building footprints, canopy height models, etc.) is straightforward, as these inputs can simply be concatenated to the low-resolution image along the channel dimension.
This could enable even higher-quality super-resolution outputs that are more faithful to the underlying land surface features, as demonstrated by~\citet{wangSemanticGuidedLarge2025}.
 The core network for image-to-image translation can also be expanded beyond a simple U-Net to more powerful architectures purpose-built for generative tasks, including Diffusion Transformers (DiT)~\citep{peeblesScalableDiffusionModels2023} and Scalable Interpolant Transformers (SiT)~\citep{maSiTExploringFlow2024}.
While not strictly necessary, perceptual losses can also be incorporated into the training procedure to provide feature-level feedback during training~\citep{berradaBoostingLatentDiffusion2024,maPixelGenImprovingPixel2026}.
Further, this approach is sensor-agnostic, and could be applied to cross-sensor super-resolution tasks (e.g., using Sentinel-2 to super-resolve Landsat 8 imagery) and multi-sensor fusion super-resolution tasks (e.g., using Sentinel-2 and Sentinel-1 to super-resolve NAIP imagery).
Ultimately, we posit that flow matching is a promising new framework for generative remote sensing super-resolution thanks to its strong generative capability, computational efficiency, ease of training, and flexibility for future extensions, addressing many of the shortcomings of existing GAN- and diffusion-based methods while opening new avenues for research and development.

\section*{Acknowledgments}

This publication is a contribution of the Mississippi Agricultural and Forestry Experiment Station (MAFES), Mississippi State University.
This work was supported by the Mississippi Space Grant Consortium (MSSGC) Graduate Fellowship Program through NASA funding, as well as by the MAFES Special Research Initiative.
We also wish to acknowledge Mississippi State University's High Performance Computing Collaboratory for providing the computational resources necessary to complete this work.

\section*{CRediT authorship contribution statement}
\textbf{Dakota Hester:} Conceptualization, Methodology, Software, Validation, Formal analysis, Investigation, Data Curation, Writing --- Original Draft, Writing --- Review \& Editing, Visualization.
\textbf{Vitor S. Martins:} Conceptualization, Methodology, Writing --- Review \& Editing, Supervision, Project administration, Funding acquisition.
\textbf{Lucas B. Ferreira:} Conceptualization, Methodology, Writing --- Review \& Editing.
\textbf{Thainara M. A. Lima:} Writing --- Review \& Editing.
\textbf{Juliana A. Araújo:} Writing --- Review \& Editing.

\bibliographystyle{elsarticle-harv}
\bibliography{bibliography}

\newpage

\section*{Supplementary material}

\renewcommand{\thesection}{S\arabic{section}}
\renewcommand{\thetable}{S\arabic{table}}
\renewcommand{\thefigure}{S\arabic{figure}}

\setcounter{section}{0}
\setcounter{table}{0}
\setcounter{figure}{0}

\begin{longtable}{@{\extracolsep{\fill}}clcl}
    \caption[
        Reclassification scheme for CBLC land use classes.
    ]{
        Reclassification scheme mapping original CBLC classes to generalized land cover classes.
    }\label{tab:cbp_reclass}
    \\
    \toprule
    \multicolumn{2}{c}{\textbf{Original CBLC Class}} & \multicolumn{2}{c}{\textbf{Generalized Class}} \\
    \cmidrule(r){1-2} \cmidrule(l){3-4}
    \textbf{Index} & \textbf{Name} & \textbf{Index} & \textbf{Name} \\
    \midrule
    \endfirsthead

    \toprule
    \multicolumn{2}{c}{\textbf{Original CBLC Class}} & \multicolumn{2}{c}{\textbf{Generalized Class}} \\
    \cmidrule(r){1-2} \cmidrule(l){3-4}
    \textbf{Index} & \textbf{Name} & \textbf{Index} & \textbf{Name} \\
    \midrule
    \endhead

    \midrule
    \multicolumn{4}{r}{{Continued on next page}} \\
    \bottomrule
    \endfoot

    \bottomrule
    \endlastfoot

    10 & Tidal Waters & \multirow{5}{*}{1} & \multirow{5}{*}{Open Water} \\
    11 & Lakes and Reservoirs & & \\
    12 & Riverine Ponds & & \\
    13 & Terrene Ponds & & \\
    14 & Streams and Rivers & & \\
    \midrule

    27 & Turf Grass & \multirow{14}{*}{2} & \multirow{14}{*}{Herbaceous} \\
    34 & Solar Field Herbaceous & & \\
    37 & Suspended Succession Herbaceous & & \\
    43 & Natural Succession Herbaceous & & \\
    46 & Harvested Forest Herbaceous & & \\
    51 & Riverine Wetlands Non-forested Herbaceous & & \\
    55 & Riverine Wetlands Harvested Forest & & \\
    61 & Terrene Wetlands Non-forested Herbaceous & & \\
    65 & Terrene Wetlands Harvested Forest & & \\
    71 & Tidal Wetlands Non-forested Herbaceous & & \\
    75 & Tidal Wetlands Harvested Forest & & \\
    81 & Cropland Herbaceous & & \\
    83 & Orchards and Vineyards Herbaceous & & \\
    86 & Pasture and Hay Herbaceous & & \\
    \midrule

    23 & Tree Canopy over Roads & \multirow{19}{*}{3} & \multirow{19}{*}{Forest Canopy} \\*
    24 & Tree Canopy over Structures & & \\*
    25 & Tree Canopy over Impervious & & \\*
    26 & Tree Canopy over Turf Grass & & \\*
    35 & Solar Field Shrubland & & \\*
    38 & Suspended Succession Shrubland & & \\*
    40 & Forest & & \\*
    41 & Forested Other & & \\*
    44 & Natural Succession Shrubland & & \\*
    52 & Riverine Wetlands Non-forested Shrubland & & \\*
    53 & Riverine Wetlands Forested Other & & \\*
    54 & Riverine Wetlands Forest & & \\*
    62 & Terrene Wetlands Non-forested Shrubland & & \\*
    63 & Terrene Wetlands Forested Other & & \\*
    64 & Terrene Wetlands Forest & & \\*
    72 & Tidal Wetlands Non-forested Shrubland & & \\*
    73 & Tidal Wetlands Forested Other & & \\*
    74 & Tidal Wetlands Forest & & \\*
    84 & Orchards and Vineyards Shrubland & & \\
    \midrule

    20 & Roads & \multirow{5}{*}{4} & \multirow{5}{*}{Impervious} \\
    21 & Structures & & \\
    22 & Other Impervious & & \\
    31 & Extractive Impervious & & \\
    32 & Solar Field Panel Arrays & & \\
    \midrule

    15 & Bare Shore & \multirow{14}{*}{5} & \multirow{14}{*}{Barren Land} \\
    28 & Bare Developed & & \\
    30 & Extractive Barren & & \\
    33 & Solar Field Barren & & \\
    36 & Suspended Succession Barren & & \\
    42 & Natural Succession Barren & & \\
    45 & Harvested Forest Barren & & \\
    50 & Riverine Wetlands Non-forested Barren & & \\
    60 & Terrene Wetlands Non-forested Barren & & \\
    70 & Tidal Wetlands Non-forested Barren & & \\
    80 & Cropland Barren & & \\
    82 & Orchards and Vineyards Barren & & \\
    85 & Pasture and Hay Barren & & \\
\end{longtable}

\end{document}